\documentclass[11pt]{article}

\usepackage[final]{acl}
\usepackage{amsmath,amssymb}
\usepackage{times}
\usepackage{latexsym}
\usepackage{booktabs}
\usepackage{multirow}
\usepackage{makecell}
\usepackage{enumitem}
\usepackage[T1]{fontenc}

\usepackage[utf8]{inputenc}

\usepackage{microtype}

\usepackage{inconsolata}

\usepackage{graphicx}

\setlist[itemize]{
	leftmargin=*,       
	itemindent=0em,     
	labelsep=0.5em,     
	listparindent=0pt,  
	parsep=0pt          
}

\title{EGTR-Review: Efficient Evidence-Grounded Scientific Peer Review Generation via Multi-Agent Teacher Distillation}

\author{
  \textbf{Xinpeng Qiu\textsuperscript{1}},
  \textbf{WANG YIHU\textsuperscript{1}},
 \textbf{Zhifeng Liu\textsuperscript{1}},
  \textbf{Xiaochen Wang\textsuperscript{1}},
  \textbf{Jimin Wang\textsuperscript{1,2,}\thanks{Corresponding author: Jiming Wang}}
\\
\\
  \textsuperscript{1}{Department of Information Management, Peking University}
\\
  \textsuperscript{2}{PKU-WUHAN Institute for Artificial Intelligence, Peking University}
\\
  \texttt{xpqiu25@stu.pku.edu.cn}\quad
  \texttt{wangyh2024@stu.pku.edu.cn}\quad
  \texttt{zhifengliu@pku.edu.cn}
  \\
  \texttt{xcwang25@stu.pku.edu.cn}\quad
  \texttt{wjm@pku.edu.cn}
}

\begin{document}
\maketitle
\begin{abstract}
Scientific peer review generation has attracted increasing attention for reducing reviewing burdens and providing timely feedback. However, existing Large Language Model (LLM)-based methods often produce generic comments with insufficient evidence support and weak source traceability, while complex multi-agent systems incur high inference costs. To address these challenges, we propose EGTR-Review, an Evidence-Grounded and Traceable Review Generation framework via Multi-Agent Teacher Distillation. EGTR-Review first constructs a multi-agent teacher that performs structure-aware paper decomposition, key-element extraction, external scholarly evidence retrieval, evidence-state labeling, verification reasoning, and review synthesis. It then distills both intermediate reasoning trajectories and final review comments into a lightweight student model through task-prefix-driven multi-task learning. An evidence-weighted objective further reduces the influence of weak, missing, or non-verifiable supervision. Experiments on public peer-review datasets show that EGTR-Review (Student) outperforms strong prompt-based, fine-tuned, and structured/agentic baselines across automatic metrics, LLM-as-Judge evaluation, and human evaluation, while maintaining strong factual grounding and source traceability with substantially lower token consumption and inference time. Our code, prompts, configurations, and sample data are available on GitHub \footnote{\url{https://anonymous.4open.science/r/EGTR-Review-3BA8}}.
\end{abstract}

\section{Introduction}

Scientific peer review is essential for research quality, but the rapid growth of submissions has increased reviewing burdens and delayed feedback \citep{https://doi.org/10.1002/asi.24810}. Automated scientific peer review generation has therefore become an important direction in AI for science \citep{darcy2024margmultiagentreviewgeneration}. Recent LLM-based methods—prompt-based, supervised fine-tuning, and multi-agent—improve the fluency and structure of generated reviews \citep{robertson2023gpt4slightlyhelpfulpeerreview, 10.1007/978-3-031-70242-6_21, 10.1016/j.inffus.2025.103332}, but still struggle in real peer-review scenarios, especially in evidence support, source traceability, paper-specific feedback, and inference efficiency.

\begin{figure}[h]
	\centering
	\includegraphics[width=1\linewidth]{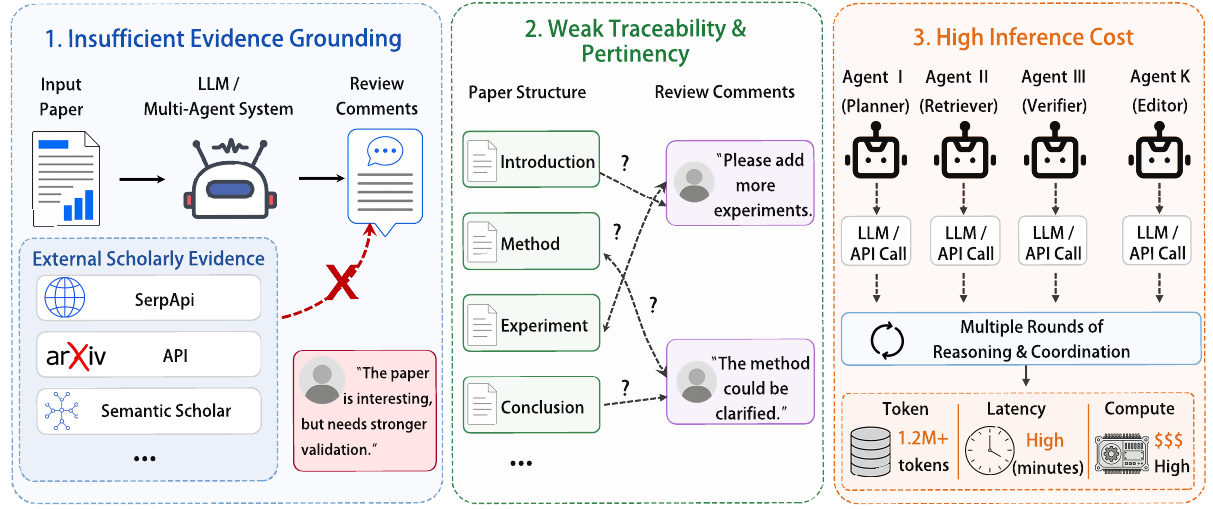}
	\caption{Key Limitations of Existing Multi-Agent Scientific Peer Review Generation Methods.}
	\label{fig1}
\end{figure}

As summarized in Figure~\ref{fig1}, existing methods face three main limitations. First, comments often lack support from paper content or external scholarly evidence, yielding generic rather than evidence-grounded feedback \citep{xu2026factreviewevidencegroundedreviewsliterature}. Second, long and complex papers make it hard to trace comments to specific sections, paragraphs, or experimental settings, weakening traceability and pertinency \citep{li-etal-2024-loogle, liu-etal-2024-lost, zhou-etal-2024-llm}. Third, multi-agent systems improve quality but rely on multi-round reasoning and repeated large-model calls, incurring high inference cost \citep{darcy2024margmultiagentreviewgeneration}. This raises a central question: how can evidence-grounded, traceable, and paper-specific review capabilities be transferred from a multi-agent teacher to a lightweight student?

In this work, we propose EGTR-Review, which builds a multi-agent teacher whose evidence-enhanced supervision integrates paper locations, scholarly evidence, evidence-state labels, reasoning trajectories, and final reviews; this is then distilled into EGTR-Review (Student). Unlike output-only distillation, the student learns the teacher's evidence-grounded reasoning process and final reviews through task-prefix-driven multi-task learning, while an evidence-weighted objective down-weights weak, missing, or non-verifiable supervision. At inference, it replaces costly teacher-side verification and synthesis but keeps paper parsing and evidence retrieval for traceability. Experiments on public peer-review datasets show that the Student outperforms strong baselines on automatic metrics, LLM-as-Judge, and human evaluation, while substantially reducing token consumption and inference time. Our main contributions are summarized as follows:

\begin{itemize}[nolistsep]
    \item We propose EGTR-Review, an evidence-grounded and traceable framework for scientific peer review generation that addresses insufficient evidence support, weak traceability, generic feedback, and high inference cost.
    \item We introduce a multi-agent teacher distillation paradigm that compresses evidence-grounded reasoning and review synthesis into an efficient student, while retaining front-end parsing and evidence retrieval for traceability. 
    \item We conduct extensive experiments on public peer-review datasets and release an anonymous reproducibility package (code, prompts, configurations, and sample distillation data), showing that EGTR-Review (Student) achieves stronger review quality and efficiency than strong baselines while maintaining factual grounding and traceability under manual verification.
\end{itemize}

EGTR-Review is intended to provide evidence-based and traceable auxiliary feedback for human reviewers, rather than replacing expert judgment in final academic decision-making.

\section{Related Works}
\subsection{Automated Scientific Peer Review Generation}
Automated scientific peer review generation aims to reduce reviewing burdens and provide timely feedback. Early studies mainly targeted supporting tasks in the peer-review workflow, such as plagiarism detection \citep{10.1145/3345317}, reviewer assignment \citep{10.1145/3292500.3330899}, and review-quality assessment \citep{zhang2022investigatingfairnessdisparitiespeer}.

With LLMs, recent work has explored full-paper review generation via prompt-based methods, supervised fine-tuning, and structured frameworks. Prompt-based methods instruct general-purpose LLMs in zero-shot or criteria-guided settings \citep{robertson2023gpt4slightlyhelpfulpeerreview, zhou-etal-2024-llm}, but often lack depth and paper-specific feedback \citep{doi:10.1056/AIoa2400196}. Supervised methods such as ReviewMT \citep{tan2024peerreviewmultiturnlongcontext} and Reviewer2 \citep{gao2024reviewer2optimizingreviewgeneration} train on paper-review pairs to improve quality. More recent structured or multi-agent systems, including $\text{SWIF}^{2}$T \citep{chamoun-etal-2024-automated}, SEA-E \citep{yu-etal-2024-automated}, TreeReview \citep{chang-etal-2025-treereview}, ScholarPeer \citep{goyal2026scholarpeercontextawaremultiagentframework}, and DeepReviewer2.0 \citep{weng2026deepreviewer20traceableagentic} decompose review generation into focused feedback, retrieval augmentation, search-grounded auditing, or traceable packaging. These demonstrate the value of structured review generation, while EGTR-Review further distills evidence-grounded reasoning trajectories and review synthesis into a lightweight student.

\subsection{Multi-Agent Reasoning Distillation}

Knowledge distillation has increasingly shifted from response-level imitation toward process-level supervision. Rationale-based and step-by-step distillation improve smaller models by transferring intermediate reasoning signals \citep{hsieh-etal-2023-distilling}, while GKD and DistiLLM extend this to autoregressive generation by aligning students with teacher behaviors and generation trajectories \citep{tan-etal-2023-gkd, 10.5555/3692070.3693067}.

Recent multi-step and multi-agent distillation, including QCRD \citep{wang-etal-2025-qcrd}, STEPER \citep{lee-etal-2025-steper}, Magdi \citep{10.5555/3692070.3692350}, AgentArk \citep{luo2026agentarkdistillingmultiagentintelligence}, and Chain-of-Agents \citep{li2025chainofagentsendtoendagentfoundation} compresses complex reasoning systems by distilling rationales, retrieval-augmented steps, interaction graphs, or collaborative agent dynamics into smaller models. However, scientific peer review generation remains underexplored here. EGTR-Review differs by distilling evidence-state-aware reasoning trajectories and final reviews from evidence-enhanced inputs grounded in paper locations, scholarly evidence, and reliability labels.

\section{Method}
\subsection{Task Setup and Framework Overview}

\begin{figure*}[h]
	\centering
	\includegraphics[width=1\linewidth]{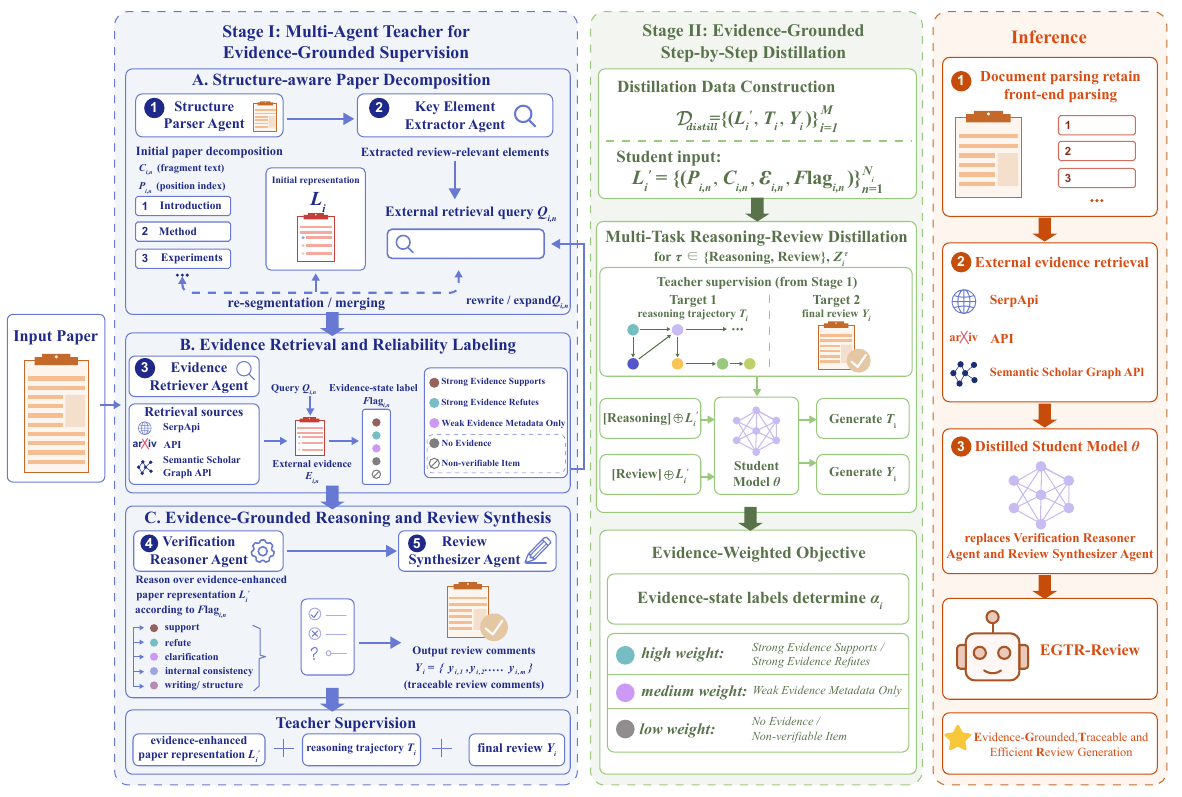}
	\caption{The overall framework of EGTR-Review.}
	\label{fig2}
\end{figure*}

Given an academic paper $D_i$, automated scientific peer review generation aims to produce review comments $Y_i=\{ y_{i,1},y_{i,2},...,y_{i,m} \}$. Unlike general long-form generation, this task requires fluency and evidence-groundedness, where judgments are grounded in paper content or external evidence; traceability, where comments link to specific sections, paragraphs, or experimental settings; pertinency, where feedback addresses methodological design, experimental validation, and argumentative weaknesses rather than generic evaluations; and efficiency, since multi-agent reasoning over long papers is costly. To meet these requirements, we propose EGTR-Review, distilled from a multi-agent teacher. As shown in Figure~\ref{fig2}, the teacher decomposes $D_i$ into position-indexed paper units, retrieves external evidence and assigns labels, then generates reasoning trajectories and final reviews. The student then takes $L_i^{'}$ as input and learns the reasoning trajectory $T_i$ and final review $Y_i$ through task prefixes. During inference, EGTR-Review retains paper parsing and evidence retrieval, while replacing costly verification reasoning and review integration with the student model.

\subsection{Multi-Agent Teacher for Evidence-Grounded Supervision}

The multi-agent teacher constructs evidence-enhanced supervision in two parts: structure-aware paper decomposition followed by evidence retrieval, reasoning, and review synthesis. EGTR-Review also applies an agent coordination mechanism between adjacent agents: the Key Element Extractor may request re-segmentation from the Structure Parser, and the Evidence Retriever may request query refinement from the Key Element Extractor when retrieval is insufficient.

\paragraph{Structure-aware Paper Decomposition.} Scientific papers have hierarchical structures, and using the entire paper $D_i$ as input makes it hard to trace comments to specific sources and strains student models with limited context windows. The teacher therefore first decomposes $D_i$ into locatable paper units.

This step is conducted by the Structure Parser Agent and the Key Element Extractor Agent. The Structure Parser Agent segments $D_i$ according to section headings, paragraph boundaries, formulas, and length thresholds, producing fragment text $C_{i,n}$ and assigning each fragment a position index $P_{i,n}$, such as its section, paragraph, figure or appendix location. The Key Element Extractor Agent then extracts review-relevant information, including research questions, methods, experimental settings, datasets, metrics, result claims, and citation evidence, and generates the retrieval query $Q_{i,n}$.

If a fragment contains multiple independent claims or lacks clear semantic boundaries, the Key Element Extractor Agent provides feedback to the Structure Parser Agent through the agent coordination mechanism for re-segmentation or merging. This feedback-based revision is performed for at most two rounds. The resulting initial representation $L_i$ contains $P_{i,n}$, $C_{i,n}$, and $Q_{i,n}$ for each paper unit.

\paragraph{Evidence Retrieval and Reliability Labeling.} Given the initial paper representation $L_i$, the Evidence Retriever Agent retrieves external scholarly evidence for each paper unit and assigns an evidence-state label. For the n-th unit of paper $D_i$, it takes the retrieval query $Q_{i,n}$ as input and searches open scholarly resources, including SerpApi, arXiv API, and Semantic Scholar Graph API. Candidate records are filtered by title matching, abstract relevance, source credibility, publication time, and semantic similarity, with at most three highly relevant records retained.

For each fragment $C_{i,n}$, the agent extracts the corresponding evidence set $\mathcal{E}_{i,n}$ and assigns an evidence-state label $Flag_{i,n}$. We use five labels: [Strong Evidence-Supports], [Strong Evidence-Refutes], [Weak Evidence-Metadata Only], [No Evidence], and [Non-verifiable Item]. The first two indicate direct support or conflict between external evidence and the paper claim. [Weak Evidence-Metadata Only]denotes weakly relevant title, abstract, or metadata evidence. [No Evidence]means no valid external evidence is retrieved, while [Non-verifiable Item]refers to subjective, structural, or writing-related fragments that cannot be directly verified through literature.

When retrieval results are insufficient or tend toward [No Evidence], the Evidence Retriever Agent uses the agent coordination mechanism to feed back the failure reason to the Key Element Extractor Agent, which rewrites or expands $Q_{i,n}$ with method names, datasets, metrics, task scenarios, or key citations within two rounds.

The evidence-enhanced representation is defined as:

\begin{equation}
	L_i^{'}=\{(P_{i,n},C_{i,n},\mathcal{E}_{i,n}, Flag_{i,n})\}_{n=1}^{N_i}
\end{equation}
where $\mathcal{E}_{i,n}$ denotes the external evidence set for the $n$-th paper unit, and $Flag_{i,n}$ denotes its evidence-state label.

\paragraph{Evidence-Grounded Reasoning and Review Synthesis.} Given the evidence-enhanced representation $L_i^{'}$, the Verification Reasoner Agent conducts evidence-state-aware reasoning for each paper unit according to $Flag_{i,n}$. For [Strong Evidence-Supports], it checks whether the retrieved evidence supports the paper’s method, experiment, or result claims. For [Strong Evidence-Refutes], it identifies conflicts between the paper claim and external evidence. For [Weak Evidence-Metadata Only], it avoids strong factual judgments and raises clarification questions. For [No Evidence], it returns to the paper content and examines internal validity, including methodological soundness, experimental sufficiency, ablation design, metric selection, argumentative consistency, and conclusion scope. For [Non-verifiable Item], it focuses on writing quality, structural organization, readability, and general peer-review criteria. These evidence-conditioned judgments form the intermediate reasoning trajectory $T_i$. A case study of evidence-grounded reasoning is provided in Appendix~\ref{appendixd1}.

The Review Synthesizer Agent then takes $L_i^{'}$ and $T_i$ as input to filter, merge, and rank candidate review points, removing redundant, weakly supported, or low-impact comments. The final review $Y_i$ keeps only comments traceable to at least one source position, evidence fragment, or reasoning step, reducing unsupported judgments. Further details are in Appendix~\ref{appendixa}.

\subsection{Evidence-Grounded Step-by-Step Distillation}

The multi-agent teacher generates well-supported and traceable review supervision through agent coordination, but its multi-round LLM calls make direct deployment costly. To improve efficiency, EGTR-Review distills the teacher-side capabilities of evidence-grounded reasoning and review synthesis into a lightweight student model. Given $L_i^{'}$, the student learns to reproduce two outputs: $T_i$ and $Y_i$. Thus, the student model distills the teacher’s reasoning and synthesis capability conditioned on $L_i^{'}$, while the Structure Parser Agent, Key Element Extractor Agent and Evidence Retriever Agent are still used to construct its structured input.

Distillation Data Construction. After constructing evidence-enhanced supervision, EGTR-Review converts the teacher outputs into distillation data. For each training paper $D_i$, the multi-agent teacher first generates the evidence-enhanced representation $L_i^{'}$, and then produces the intermediate reasoning trajectory $T_i$ and final review comments $Y_i$. The distillation dataset is defined as:

\begin{equation}
	D_{distill}=\{(L_i^{'},T_i,Y_i ) \}_{i=1}^{M}
\end{equation}
where $M$ denotes the number of training papers, $L_i^{'}$ contains source positions, paper fragments, external evidence, and evidence-state labels, $T_i$ denotes the teacher-side verification reasoning trajectory under different evidence states, and $Y_i$ denotes the final review comments. This dataset provides both process supervision through $T_i$ and outcome supervision through $Y_i$, enabling the student to learn how review judgments are formed and how final reviews are generated.

\paragraph{Multi-Task Reasoning-Review Distillation.} We formulate distillation as a task-prefix-driven multi-task generation problem. For the same evidence-enhanced representation $L_i^{'}$, the student learns two tasks with different prefixes: [Reasoning], which guides the model to generate the teacher-side reasoning trajectory $T_i$, and [Review], which guides it to generate the final review $Y_i$. For each task $\tau \in \{Reasoning,Review\}$, let $Z_i^{\tau}$ denote the target sequence, where $Z_i^{Reasoning}=T_i$ and $Z_i^{Review}=Y_i$. Given the prefixed input $[ \tau ] \oplus L_i^{'}$, the student generates $Z_i^{\tau}$ autoregressively and minimizes the negative log-likelihood:

\begin{equation}
	L_i^{\tau} = -\frac{1}{|Z_i^{\tau}|} \sum_{t=1}^{|Z_i^{\tau}|} \log p_\theta \left( Z_{i,t}^{\tau} \mid Z_{i,<t}^{\tau}, [\tau] \oplus L_i' \right)
\end{equation}
where $[\tau]$ is the task prefix, $\oplus$ denotes concatenation, $Z_{i,t}^{\tau}$ denotes the $t$-th target token, $Z_{i,<t}^{\tau}$ denotes the sequence of previous target tokens before position $t$, and $\theta$ denotes the student parameters.

\paragraph{Evidence-Weighted Objective.} Evidence reliability varies across paper units. Treating all supervision signals equally may cause the student to overfit unstable judgments from weak-evidence, no-evidence, or non-verifiable fragments. Therefore, EGTR-Review assigns training weights according to evidence-state labels. Units with [Strong Evidence-Supports]or [Strong Evidence-Refutes]receive higher weights, while units with [Weak Evidence-Metadata Only], [No Evidence], or [Non-verifiable Item]receive lower weights. The unit-level weights are then aggregated within each paper to obtain a sample-level weight $ \alpha_i$.

The final training objective is defined as:

\begin{equation}
	\mathcal{L} = \frac{1}{M} \sum_{i=1}^{M} \alpha_i  (\mathcal{L}_i^{Reasoning} + \lambda \mathcal{L}_i^{Review}  )
\end{equation}
where $M$ is the number of training papers, $\mathcal{L}_i^{Reasoning}$ and $\mathcal{L}_i^{Review}$ denote the reasoning and review generation losses for paper $D_i$, $\alpha_i$ adjusts the contribution of $D_i$ according to evidence reliability, and $\lambda$ balances reasoning supervision and review supervision.

During inference, EGTR-Review retains document parsing and evidence retrieval, while using the student model to replace the teacher-side Verification Reasoner Agent and Review Synthesizer Agent. More details of EGTR-Review are provided in Appendix~\ref{appendixa}.

\section{Experiments}
\subsection{Experimental Settings}
\paragraph{Datasets.} We build our benchmark from two public peer-review resources, PeerRead and OpenReview, covering ICLR 2017–2024 submissions with their reviews. After preprocessing and filtering, it contains 1,386 papers (997 training, 60 validation, and 329 test). The test set is approximately balanced between accepted and rejected submissions and spans diverse computer science topics. The training set builds teacher supervision, the validation set is for hyperparameter selection and early stopping, and the test set only for final evaluation. To prevent review leakage, retrieval and teacher-side supervision exclude human reviews, rebuttals, decisions, ratings, and confidence scores. More details on dataset are in Appendix~\ref{appendixb1}, and the case study is in Appendix~\ref{appendixd}.

\paragraph{Baselines.} We compare EGTR-Review against three groups of baselines. The first is prompt-based: Zero-Shot LLM \citep{robertson2023gpt4slightlyhelpfulpeerreview}, which generates reviews with zero-shot prompting, and Criteria-guided LLM \citep{zhou-etal-2024-llm}, which adds explicit review dimensions to the prompt. The second is supervised methods, ReviewMT-SFT \citep{tan2024peerreviewmultiturnlongcontext} and Reviewer2 \citep{gao2024reviewer2optimizingreviewgeneration}, examining the effect of fine-tuning on peer-review data. The third is structured or agentic methods, $\text{SWIF}^{2}$T \citep{chamoun-etal-2024-automated}, SEA-E \citep{yu-etal-2024-automated}, and TreeReview \citep{chang-etal-2025-treereview}, based on task decomposition, retrieval augmentation, or structured reasoning. We also report EGTR-Review (Teacher) and EGTR-Review (Student) to compare the full teacher with the distilled student. Appendix~\ref{appendixb} gives additional baseline details.

\paragraph{Implementation Details.} For supervised review generation baselines whose released checkpoints may overlap with our ICLR benchmark, we re-implement or adapt them using the same splits as EGTR-Review. For API-based and structured agentic baselines, we use GPT-5.1 as the backbone LLM and keep prompts, decoding parameters, maximum output length, and token accounting consistent. We provide model access settings, prompt templates, configurations, representative input-output logs, and reproduction scripts in the anonymous repository. EGTR-Review is not tied to GPT-5.1 and can be instantiated with other long-context instruction-following LLMs. Student training is conducted on two NVIDIA A100 GPUs. More details are provided in Appendix~\ref{appendixb}.

\paragraph{Hyperparameters Setup.} For all API-based baselines and EGTR-Review (Teacher), we set the temperature to 0 and the maximum decoding length to 8,192 tokens per LLM call. For student distillation, we use Qwen2.5-7B-Instruct and train it for 3 epochs with a maximum sequence length of 4,096 per distillation instance. We set the learning rate to $2.0 \times 10^{-4}$, batch size to 1, gradient accumulation steps to 32, warmup ratio to 0.05, and use a cosine learning-rate scheduler. We enable bf16 and gradient checkpointing, set the maximum gradient norm to 1.0, and fix the random seed to 42. In the efficiency analysis, token usage is counted cumulatively per paper across all inference calls and intermediate stages.

\begin{table*}[t]
	\centering
	\caption{Automatic Evaluation Results on Review Generation Quality.}
	\label{tab1}
	\small
	\renewcommand{\arraystretch}{1.15}
	\setlength{\tabcolsep}{5pt}
	\begin{tabular}{lcccccccc}
		\toprule
		\multirow{2}{*}{\textbf{Method}} 
		& \multicolumn{4}{c}{\textbf{Lexical and Semantic Similarity}} 
		& \multicolumn{4}{c}{\textbf{Semantic Alignment and Specificity}} \\
		\cmidrule(lr){2-5} \cmidrule(lr){6-9}
		& \textbf{R-1} 
		& \textbf{R-2} 
		& \textbf{R-L} 
		& \textbf{BERTScore} 
		& \textbf{SN-P} 
		& \textbf{SN-R} 
		& \textbf{SN-F1} 
		& \textbf{ITF-IDF} \\
		\midrule
		
		Zero-Shot LLM 
		& 30.42 & 6.15 & 14.85 & 81.16 
		& 30.85 & 33.20 & 31.99 & 3.68 \\
		
		Criteria-guided LLM 
		& 33.68 & 6.58 & 16.79 & 82.18 
		& 33.40 & 35.65 & 34.49 & 3.79 \\
		
		ReviewMT-SFT 
		& 35.76 & 8.75 & 18.75 & 82.80 
		& 34.20 & 36.50 & 35.31 & 3.73 \\
		
		Reviewer2 
		& 36.88 & 8.04 & 17.77 & 83.65 
		& 36.40 & 38.10 & 37.23 & 3.91 \\
		
		$\text{SWIF}^{2}$T 
		& 41.20 & 13.17 & 21.30 & 83.70 
		& 40.50 & 42.20 & 41.33 & 4.08 \\
		
		SEA-E 
		& 42.80 & 12.90 & 22.30 & 83.65 
		& 42.10 & 44.00 & 43.03 & 4.26 \\
		
		TreeReview 
		& 48.90 & 14.15 & 22.75 & 84.45 
		& 46.80 & 48.96 & 47.86 & 4.78 \\
		
		EGTR-Review (Student) 
		& 49.20 & 14.35 & 23.60 & 85.60 
		& 47.44 & 49.51 & 48.45 & 5.14 \\
		
		\textbf{EGTR-Review (Teacher)} 
		& \textbf{55.91} & \textbf{16.30} & \textbf{26.80} & \textbf{87.40} 
		& \textbf{53.60} & \textbf{56.20} & \textbf{54.87} & \textbf{5.47} \\
		
		\bottomrule
	\end{tabular}
\end{table*}

\subsection{Main Results}\label{sec4.2}

We evaluate EGTR-Review from three perspectives: automatic metrics, LLM-as-Judge, and human evaluation. Since peer review comments are open-ended, a single text-similarity metric cannot fully capture review quality, so we use ROUGE \citep{lin-2004-rouge}, BERTScore \citep{zhang2020bertscoreevaluatingtextgeneration}, SN-F1 \citep{lou2025aaar10assessingaispotential}, ITF-IDF \citep{du-etal-2024-llms}, LLM-as-Judge scores, and expert preference. Appendix~\ref{appendixc} gives metric definitions and human-evaluation protocols, and Appendix~\ref{appendixe} gives the LLM-as-Judge prompts and output formats.

Table~\ref{tab1} reports automatic evaluation results across lexical similarity, semantic similarity, semantic alignment, and paper specificity. EGTR-Review (Student) achieves the best performance among all external baselines, with 23.60 R-L, 85.60 BERTScore, 48.45 SN-F1, and 5.14 ITF-IDF. Compared with the strongest structured baseline, TreeReview, the Student improves R-L and BERTScore by 0.85 and 1.15 points, respectively, and further improves SN-F1 and ITF-IDF by 1.2\% and 7.5\%. These consistent gains suggest that evidence-enhanced inputs and reasoning-trajectory distillation help the student generate reviews that are not only closer to reference reviews, but also more semantically aligned and paper-specific. EGTR-Review (Teacher) achieves the highest scores across all metrics, reaching 26.80 R-L, 87.40 BERTScore, 54.87 SN-F1, and 5.47 ITF-IDF, indicating the upper-bound performance of the full multi-agent system. Overall, the Student retains much of the teacher’s review-generation capability while outperforming strong prompt-based, fine-tuned, and structured/agentic baselines. Additional details are provided in Appendix~\ref{appendixc1}.

\paragraph{LLM-as-Judge and Human Evaluation.} We adopt LLM-as-Judge for multi-dimensional assessment over Pertinency, Usefulness, Evidence-groundedness, Traceability, Depth of Analysis, Comprehensiveness, and Overall. All dimensions are scored on a 1–10 scale.To reduce single-judge bias, we cross-check with multiple heterogeneous strong models. The complete evaluation prompts and output format are provided in Appendix~\ref{appendixe}.

\begin{figure}
	\centering
	\includegraphics[width=1\linewidth]{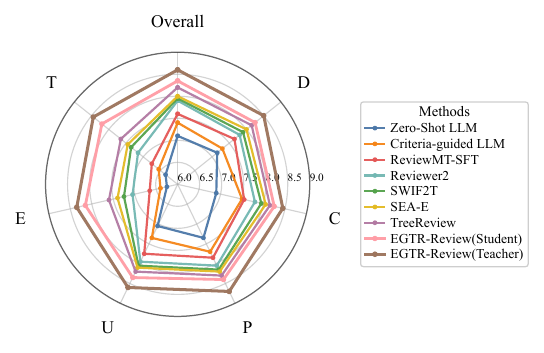}
	\caption{LLM-as-Judge Evaluation across Review Quality Dimensions. T, E, U, P, C, and D denote Traceability, Evidence-groundedness, Usefulness, Pertinency, Comprehensiveness, and Depth of Analysis, respectively.}
	\label{fig3}
\end{figure}

Figure~\ref{fig3} reports the LLM-as-Judge results across seven review quality dimensions. EGTR-Review (Teacher) achieves the best performance on all dimensions, reaching Overall 8.60 ($\uparrow$4.9\% over TreeReview), with the largest gains on Traceability 8.45 ($\uparrow$10.5\%) and Evidence-groundedness 8.35 ($\uparrow$9.9\%). EGTR-Review (Student) also outperforms all external baselines, achieving Overall 8.35 ($\uparrow$1.8\% over TreeReview), with clear advantages in Evidence-groundedness 8.15 ($\uparrow$7.2\%) and Traceability 8.20 ($\uparrow$7.2\%). Although below the Teacher, the Student surpasses TreeReview on Pertinency, Depth of Analysis, and Comprehensiveness, showing that distillation preserves much of the teacher’s evidence-aware and position-sensitive reviewing capability while improving efficiency.

\begin{figure}[h]
	\centering
	\includegraphics[width=1\linewidth]{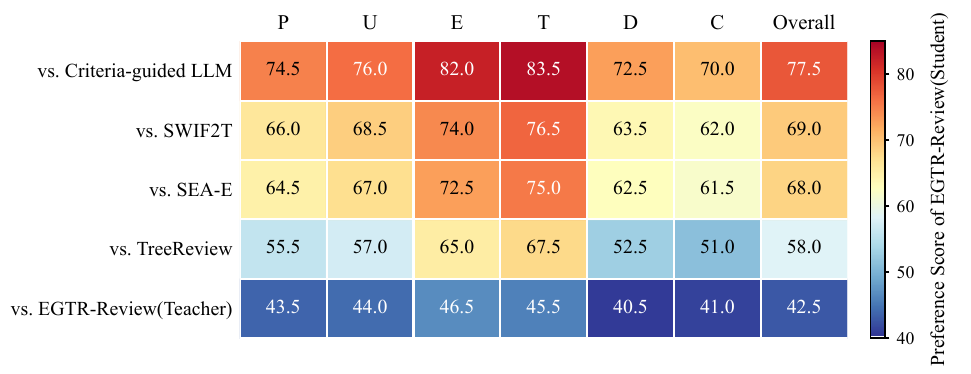}
	\caption{Human Preference Evaluation of EGTR-Review.}
	\label{fig4}
\end{figure}

We conduct a human preference evaluation using anonymous pairwise comparison. We sample 30 test papers and recruit five expert peer-review evaluators. Each questionnaire presents the paper and two anonymized reviews in randomized order; evaluators select the better review or mark a tie. Figure~\ref{fig4} reports the preference results. EGTR-Review (Student) is preferred over all external baselines, with Overall scores: 77.5 vs. Criteria-guided LLM, 69.0 vs. $\text{SWIF}^{2}$T, 68.0 vs. SEA-E, and 58.0 vs. TreeReview. Its largest advantages over TreeReview appear in Evidence-groundedness and Traceability (65.0 and 67.5), confirming the benefit of evidence retrieval, evidence-state labels, and position-indexed paper units. Against EGTR-Review (Teacher), the Student scores 42.5, confirming the full teacher remains the quality upper bound. Additional details are provided in Appendix~\ref{appendixc2} and \ref{appendixc3}.

\subsection{Evidence Grounding and Traceability Evaluation}

Beyond overall quality, we assess factual support and traceability on 50 randomly sampled test papers annotated by researchers with peer-review experience. FActScore measures the proportion of atomic factual claims supported by the source paper or retrieved evidence \citep{min-etal-2023-factscore}, while Traceability Accuracy (TA) measures the proportion of comments linkable to specific paper units, experimental settings, or evidence records \citep{chamoun-etal-2024-automated}. Appendix~\ref{appendixc4} provides the full annotation procedures.

\begin{table}[htbp]
	\centering
	\caption{Manual Factual Grounding and Traceability Evaluation.}
	\label{tab2}
	\small
	\renewcommand{\arraystretch}{1.15}
	\begin{tabular}{lcc}
		\toprule
		\textbf{Model} & \textbf{FActScore} & \textbf{TA} \\
		\midrule
		EGTR-Review (Student) & 0.746 & 0.812 \\
		EGTR-Review (Teacher) & \textbf{0.823} & \textbf{0.904} \\
		\bottomrule
	\end{tabular}
\end{table}

\begin{table*}[t]
	\centering
	\caption{Ablation Results across Automatic Metrics and LLM-as-Judge Evaluation. P, E, T, and D denote Pertinency, Evidence-groundedness, Traceability, and Depth of Analysis. Scores are averaged over Gemini-3.1-Pro-Preview, DeepSeek-V4, and Claude-Opus-4.6.}
	\label{tab3}
	\footnotesize
	\renewcommand{\arraystretch}{1.15}
	\setlength{\tabcolsep}{4pt}
	\begin{tabular}{@{}>{\raggedright\arraybackslash}p{2.1cm}
			>{\raggedright\arraybackslash}p{5.0cm}
			cccccccc@{}}
		\toprule
		\multirow{2}{*}{\textbf{Category}} 
		& \multirow{2}{*}{\textbf{Method}} 
		& \multicolumn{3}{c}{\textbf{Automatic Metrics}} 
		& \multicolumn{5}{c}{\textbf{LLM-as-Judge Dimensions}} \\
		\cmidrule(lr){3-5} \cmidrule(lr){6-10}
		& & \textbf{R-L} & \textbf{BERTScore} & \textbf{SN-F1} 
		& \textbf{P} & \textbf{E} & \textbf{T} & \textbf{D} & \textbf{Overall} \\
		\midrule
		
		\multirow{5}{*}{Teacher-side}
		& w/o structure parser 
		& 24.65 & 86.10 & 52.35 
		& 8.15 & 8.30 & 7.45 & 8.12 & 8.05 \\
		
		& w/o key-element extraction 
		& 24.20 & 85.92 & 51.78 
		& 8.06 & 8.18 & 7.90 & 8.03 & 7.98 \\
		
		& w/o agent coordination 
		& 23.85 & 85.70 & 50.95 
		& 7.92 & 7.86 & 7.62 & 7.88 & 7.82 \\
		
		& w/o evidence retrieval \& reliability labeling 
		& 23.10 & 85.05 & 49.60 
		& 7.72 & 6.88 & 7.30 & 7.64 & 7.50 \\
		
		& EGTR-Review (Teacher) 
		& \textbf{26.80} & \textbf{87.40} & \textbf{54.87} 
		& \textbf{8.62} & \textbf{8.35} & \textbf{8.45} & \textbf{8.54} & \textbf{8.60} \\
		
		\midrule
		
		\multirow{5}{*}{Student-side}
		& w/o reasoning supervision 
		& 21.05 & 84.20 & 41.10 
		& 7.70 & 7.42 & 7.55 & 7.48 & 7.58 \\
		
		& w/o task-prefix multi-task learning 
		& 20.40 & 83.85 & 37.60 
		& 7.48 & 7.20 & 7.32 & 7.25 & 7.35 \\
		
		& w/o evidence labels 
		& 20.75 & 84.00 & 38.40 
		& 7.51 & 7.16 & 7.38 & 7.28 & 7.36 \\
		
		& w/o evidence-weighted objective 
		& 22.85 & 85.20 & 46.90 
		& 8.05 & 7.82 & 8.00 & 7.95 & 8.03 \\
		
		& EGTR-Review (Student) 
		& \textbf{23.60} & \textbf{85.60} & \textbf{48.45} 
		& \textbf{8.31} & \textbf{8.15} & \textbf{8.20} & \textbf{8.18} & \textbf{8.35} \\
		
		\bottomrule
	\end{tabular}
\end{table*}

Table~\ref{tab2} reports the evidence-quality and traceability results. EGTR-Review (Teacher) outperforms EGTR-Review (Student), reaching 0.823 on FActScore and 0.904 on TA, indicating that the full multi-agent teacher produces more factually supported and traceable review comments. EGTR-Review (Student) still attains 0.746 on FActScore and 0.812 on TA, showing that evidence-grounded distillation preserves much of the teacher's factual grounding and source-tracing ability. The remaining gap suggests that precise evidence verification and comment-to-source alignment are still better handled by the full teacher system.

\subsection{Ablation Studies}

To assess each component's contribution, we ablate four teacher-side modules (structure parser, key-element extractor, agent coordination, and evidence retrieval and reliability labeling) and four student-side components (reasoning-trajectory supervision, task-prefix multi-task learning, evidence-state labels, and the evidence-weighted objective). Table~\ref{tab3} reports both automatic-evaluation and LLM-as-Judge results.

Both teacher-side supervision and student-side distillation contribute to EGTR-Review. On the teacher side, removing evidence retrieval and reliability labeling causes the largest degradation (Evidence-groundedness drops from 8.35 to 6.88), confirming the role of external evidence and evidence-state labels in producing verifiable reviews; removing agent coordination weakens performance, indicating that inter-agent feedback helps performance. On the student side, removing evidence labels reduces evidence-groundedness and traceability and removing the evidence-weighted objective causes a milder drop, whereas removing task-prefix multi-task learning leads to strongest decline (Overall drops from 8.35 to 7.35), underscoring the value of separating reasoning and review-generation objectives.

\subsection{Efficiency and Cost Analysis}

We compare the inference cost and efficiency of review generation systems. Table~\ref{tab4} shows that EGTR-Review (Student) is more efficient than the full teacher and other agentic baselines. Compared with EGTR-Review (Teacher), the student reduces total tokens per paper from 308,800 to 105,124 and inference time from 139s to 44s, while retaining evidence-grounded review capability. Although EGTR-Review (SFT) is lighter, it removes evidence-enhanced reasoning. Distillation reduces deployment cost while preserving stronger review-generation ability than purely lightweight variants.

\begin{table}[htbp]
	\centering
	\caption{Cost comparison of different review generation systems.}
	\label{tab4}
	\small
	\renewcommand{\arraystretch}{1.15}
	\resizebox{\columnwidth}{!}{
		\begin{tabular}{lcccc}
			\toprule
			\textbf{Method} 
			& \makecell{\textbf{Input}\\\textbf{Tokens/Paper}} 
			& \makecell{\textbf{Output}\\\textbf{Tokens/Paper}} 
			& \makecell{\textbf{Total}\\\textbf{Tokens/Paper}} 
			& \makecell{\textbf{Inference}\\\textbf{Time/Paper}} \\
			\midrule
			$\text{SWIF}^{2}$T & 196,850 & 21,800 & 218,650 & 82s \\
			TreeReview & 407,600 & 38,850 & 446,450 & 172s \\
			EGTR-Review (SFT) & 17,800 & 1,520 & 19,320 & 10s \\
			EGTR-Review (Teacher) & 278,600 & 30,200 & 308,800 & 139s \\
			EGTR-Review (Student) & 96,700 & 8,424 & 105,124 & 44s \\
			\bottomrule
		\end{tabular}
	}
\end{table}

\section{Conclusion}

In this paper, we presented EGTR-Review, an evidence-grounded and traceable framework for scientific peer review generation via multi-agent teacher distillation. The teacher decomposes papers, retrieves scholarly evidence, assigns evidence-state labels, and conducts verification reasoning, while the student distills its reasoning trajectories and final comments from evidence-enhanced inputs, reducing deployment cost. Experiments show gains in review quality, evidence-groundedness, traceability, and efficiency over strong baselines. Consistent with prior work, EGTR-Review is intended to assist human reviewers rather than automate academic decisions. Future work will improve retrieval reliability and human-in-the-loop verification.

\section*{Limitations}

Despite the encouraging results achieved by EGTR-Review, this work still has two main limitations that should be addressed in future work:

\paragraph{Domain Generalization.} EGTR-Review is mainly evaluated in AI-related peer-review scenarios, including papers from areas such as machine learning, natural language processing, and computer vision. Although these settings cover a broad range of current AI conference submissions, review criteria and evidential expectations may differ across scientific domains. Future work can further evaluate EGTR-Review in additional fields such as biomedicine, where experimental validation, domain-specific evidence, and safety-related concerns may require different review standards.

\paragraph{Multimodal Review Consideration.} The current framework mainly focuses on text-based paper content, structured evidence, and traceable review comment generation. However, scientific peer review often also involves assessing visual and multimodal presentation, such as whether figures and tables are clearly organized, whether captions are sufficiently informative, whether visual evidence supports the corresponding claims, and whether the text, tables, and figures are mutually consistent. Future work can extend EGTR-Review toward multimodal review assistance, enabling the system to evaluate the readability, standardization, and text-figure consistency of tables, figures, and experimental visualizations, and to provide more specific suggestions for improving the presentation and interpretability of scientific papers.

\section*{Ethical Considerations}

EGTR-Review is intended to assist human reviewers rather than replace experts in final academic judgments. Automatically generated reviews may inherit biases from training data, retrieval resources, and underlying language models. If directly used for formal review decisions, such outputs may introduce unfair or misleading risks. Therefore, system outputs should be treated as review-assistance materials that help reviewers identify potential issues, inspect evidence clues, and improve review efficiency, but they should not serve as the sole basis for paper acceptance, rejection, or academic evaluation. 

To reduce ethical risks, we filter OpenReview reviews, author rebuttals, accept/reject decisions, and reference reviews already included in the dataset during external retrieval. However, indirect exposure to derivative discussions or metadata correlated with review outcomes cannot be fully excluded. The system retains evidence sources, original paper locations, and evidence-state labels, helping users trace the basis of review comments. For low-confidence cases such as Weak Evidence-Metadata Only, No Evidence, and Non-verifiable Item, the system generates cautious questions or internal consistency checks rather than strong factual judgments.

\section*{Acknowledgments}
This work was partially supported by the research project "Research on Methods for Detecting Artificial Intelligence-Generated Content" (No. 8430503406). We would like to thank all the anonymous reviewers for their valuable comments and constructive feedback.

\bibliography{custom}

\appendix

\section{More details of EGTR-Review}
\label{appendixa}
\subsection{Agent Coordination Mechanism Details}\label{appendixa1}
EGTR-Review introduces an agent coordination mechanism within the multi-agent teacher to improve the stability of teacher-side supervision signals. Here, agent coordination means that downstream agents provide targeted feedback to upstream agents at locally unstable points, rather than performing open-ended multi-round debate or relying on a complex communication framework. Its goal is to reduce the impact of segmentation errors and under-specified queries, without changing the overall pipeline for constructing $L_i^{'}$, $T_i$, and $Y_i$ described in the main text.

\paragraph{Segment Adjustment.} During structure-aware decomposition, the system checks whether each paper unit is suitable for subsequent evidence retrieval and review generation. If a segment is overly long, contains multiple independent claims, or becomes semantically incomplete due to improper boundaries, the Key Element Extractor Agent provides feedback to the Structure Parser Agent through the agent coordination mechanism for re-segmentation or merging. This process only revises clearly problematic segments, rather than re-parsing the entire paper, and is performed for at most two rounds.

\paragraph{Query Refinement.} During evidence retrieval, the system performs limited query refinement according to the quality of retrieved results. When candidate results are overly broad, weakly related to the paper segment, or insufficient for assigning a reliable evidence state, the Evidence Retriever Agent uses the agent coordination mechanism to feed back the failure reason to the Key Element Extractor Agent, which rewrites or expands $Q_{i,n}$ with method names, datasets, task scenarios, evaluation metrics, key terms, or citation information within two rounds. This step reduces evidence missingness caused by under-specified query expressions.

\subsection{Evidence Labeling and Reasoning Details}\label{appendixa2}
EGTR-Review uses five evidence status labels to guide the reasoning path of the Verification Reasoner Agent.

\begin{itemize}[noitemsep]
	\item \textbf{Strong Evidence-Supports:} For [Strong Evidence-Supports], external evidence supports the method, experiment, or result claim in the paper segment. The system checks whether the paper uses this evidence properly and whether the claim is over-stated, under-cited, or lacks clear scope.
	\item \textbf{Strong Evidence-Refutes:} For [Strong Evidence-Refutes], the system checks whether the paper’s claim stays within the scope of the retrieved evidence and avoids over-generalization or unclear conclusion boundaries.
	\item \textbf{Weak Evidence-Metadata Only:} For [Weak Evidence-Metadata Only], only weak metadata-level information is available, such as titles, abstracts, years, venues, or citation relations. The system therefore avoids strong factual judgments and generates cautious clarification questions.
	\item \textbf{No Evidence:} For [No Evidence], no reliable external evidence is retrieved. The unit is not discarded; instead, the system turns to paper-internal grounding, checking methodological validity, experimental sufficiency, ablation design, metric selection, argument consistency, and conclusion boundaries.
	\item \textbf{Non-verifiable Item:} For [Non-verifiable Item], the segment mainly concerns writing, organization, or subjective evaluation. The system does not make external factual judgments and only keeps comments related to clarity, structure, readability, or review criteria.
	\item \textbf{Overall:} EGTR-Review requires each comment to be grounded in locatable paper content and an explicit reasoning process. When external evidence is available, it is further incorporated into the verification chain. Detailed prompts are provided in Appendix~\ref{appendixe}.
\end{itemize}

\section{Experimental Setting Details}
\label{appendixb}
\subsection{Benchmark Construction}\label{appendixb1}
We construct our experimental corpus from PeerRead and OpenReview. To keep the domain and review format consistent, we retain publicly available ICLR submissions from 2017 to 2024 with their corresponding reviews. We remove samples with missing paper text, missing reviews, duplicate records, or unmatched paper–review pairs. The final corpus contains 1,386 papers and their reviews.

We split the corpus into 997 training papers, 60 validation papers, and 329 test papers. The training set is used to run multi-agent teachers and construct $D_{distill}$. The validation set is used for hyperparameter selection and early stopping. The test set is used only for final evaluation. The corpus mainly covers common ICLR topics, such as reinforcement learning, machine learning, representation learning, optimization, and generative modeling.

To reduce data leakage, all splits are performed at the paper level. During teacher-side supervision construction, multi-agent teachers can only access the initial paper text, structural information, and external scholarly evidence, but not human reviews, author responses, decisions, ratings, or confidence scores. During evidence retrieval, we exclude OpenReview pages or fields related to reviews, rebuttals, decisions, ratings, and confidence scores, and only retain papers, abstracts, metadata, citation information, and other scholarly materials as evidence sources. The test set is never used for model selection, prompt tuning, distillation data construction, or student training.

\begin{figure}
	\centering
	\includegraphics[width=1\linewidth]{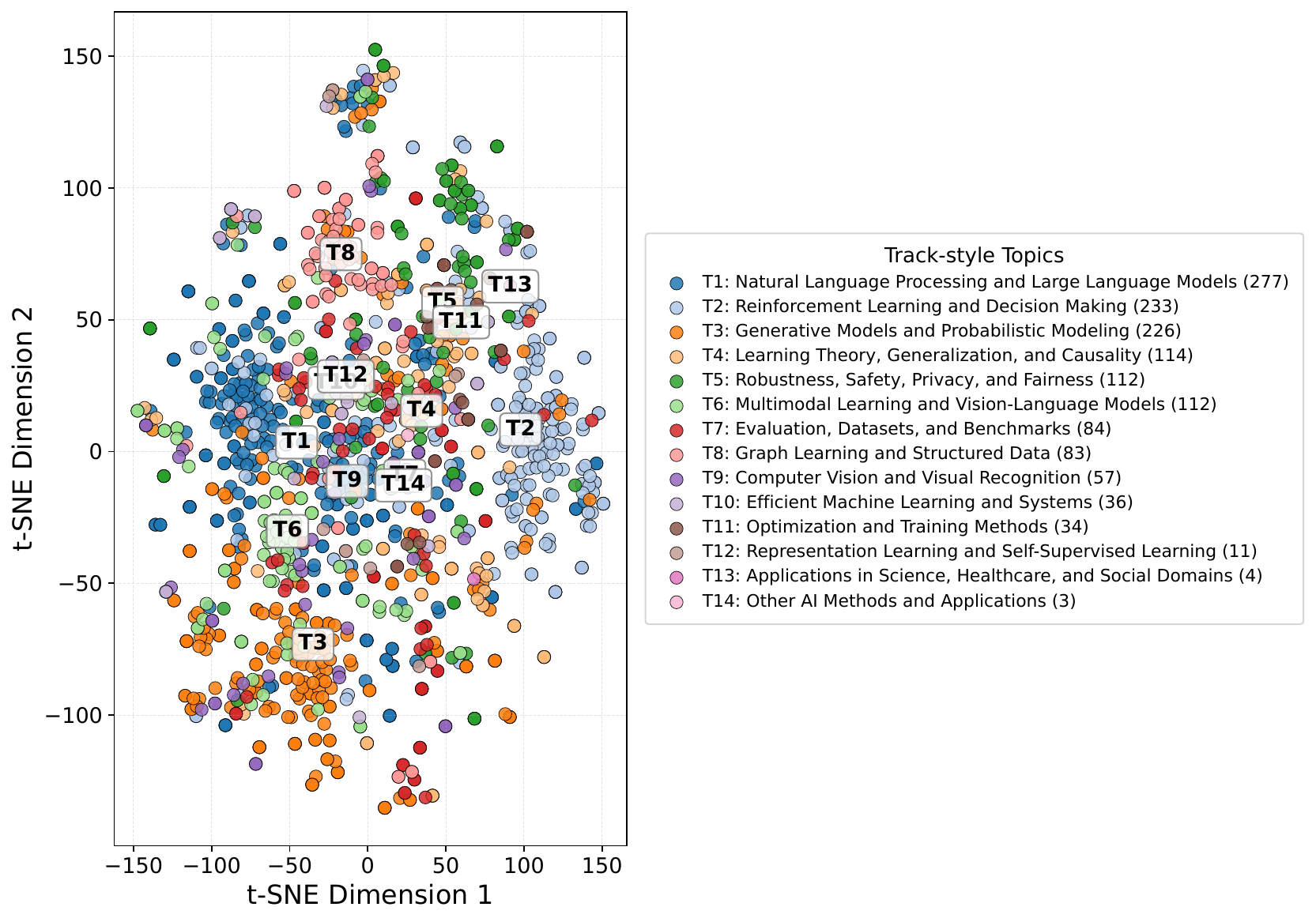}
	\caption{t-SNE Visualization of Topic Distribution in the Full Corpus.}
	\label{fig5}
\end{figure}

The t-SNE visualization shows that the collected papers are distributed across multiple topic regions rather than being concentrated in a single narrow area. The corpus covers a broad range of computer science topics with a focus on AI and machine learning, including natural language processing and large language models, reinforcement learning and decision making, generative models, learning theory, robustness and safety, multimodal learning, graph learning, efficient machine learning, and evaluation-oriented studies. This distribution suggests that the benchmark contains diverse research themes and review concerns, which helps reduce the risk that the evaluation is dominated by a small set of homogeneous topics. For the final evaluation split, we further maintain an approximately balanced distribution of accepted and rejected submissions to reduce potential decision-distribution bias. Overall, the dataset provides a reasonably diverse and balanced evaluation setting for assessing scientific peer review generation methods.

\subsection{Baseline Implementation Details}\label{appendixb2}
\paragraph{Zero-Shot LLM.}  This baseline directly calls GPT-5.1 to generate review comments, without task-specific fine-tuning, external retrieval, multi-agent collaboration, or distillation \citep{robertson2023gpt4slightlyhelpfulpeerreview}. In implementation, we input the full paper into GPT-5.1 and provide a unified review prompt, requiring the model to generate structured review comments, including summary, strengths, weaknesses, and questions. This baseline relies only on the contextual understanding and generation capability of GPT-5.1, without using $L_i^{'}$, $Flag_{i,n}$, $T_i$, or the evidence-weighted objective. For fair comparison, all samples use the same prompt template, and the temperature, maximum output length, and other inference settings are kept consistent with those of other API-based baselines. This method is used to measure the basic performance of a general-purpose LLM in generating scientific peer reviews under the setting of no retrieval, no distillation, and no explicit evidence-based reasoning supervision.

\paragraph{Criteria-guided LLM.} This baseline is used to evaluate the effect of explicit review-dimension constraints on LLM-based peer review generation. Following the review generation setting of \citet{zhou-etal-2024-llm}, this method provides several review criteria in the prompt and guides the model to generate an ICLR-style review from aspects such as summary, motivation, substance, originality, soundness, clarity, replicability, and meaningful comparison. Unlike Zero-Shot LLM, this baseline not only asks the model to directly read the paper and generate a review, but also explicitly specifies the review dimensions and output structure in the prompt to improve the coverage of generated comments. In this paper, we implement it as a prompt-based baseline: the input is the full paper, the underlying model is uniformly set to GPT-5.1, and the output includes summary, strengths, weaknesses, and questions. This baseline does not use external evidence retrieval, evidence-state labels, multi-agent teachers, or distillation training; it only relies on criteria-guided prompting to constrain the generation process of GPT-5.1. For fair comparison, the temperature, maximum output length, and other API inference parameters are kept consistent with those of other API-based baselines.

\paragraph{ReviewMT-SFT.} ReviewMT \citep{tan2024peerreviewmultiturnlongcontext} formulates peer review as a multi-turn dialogue involving a reviewer, an author, and a decision maker, and constructs multi-turn reviewing data from public resources such as OpenReview. In our experiments, we adapt its supervised fine-tuning setting to the single-turn review generation task. Specifically, we use the paper-review pairs in our training set to construct SFT data, where the input consists of the paper title, abstract, and main text, and the output is the corresponding human review organized into structured fields, including summary, strengths, weaknesses, and questions. This baseline only learns the mapping from paper text to final review text, without external evidence retrieval, evidence status labels, multi-agent teachers, reasoning traces, or evidence-weighted objective. We follow the SFT setting of ReviewMT and train the model on the same train/validation/test split as EGTR-Review (Student) for fair comparison. This baseline is used to evaluate response-only supervised fine-tuning against our method, which jointly distills evidence-enhanced inputs and intermediate reasoning traces.

\paragraph{REVIEWER2.} REVIEWER2 \citep{gao2024reviewer2optimizingreviewgeneration} is a two-stage SFT-based review generation method that explicitly controls the aspects covered by generated reviews. It first fine-tunes a prompt generation model $M_p$ to produce aspect prompts from the full paper, and then fine-tunes a review generation model $M_r$ to generate the final review conditioned on both the paper and the generated aspect prompt. Following this framework, we re-implement REVIEWER2 on our own train/validation/test split to avoid potential data leakage from its released weights, since our benchmark focuses on ICLR 2017–2024 papers. Specifically, we first construct aspect prompts for training reviews following the PGE-style procedure, then train $M_p$ to map each paper to aspect prompts and $M_r$ to map the paper–aspect pair to the corresponding structured review. During inference, $M_p$ generates an aspect prompt for each test paper, and $M_r$ produces the final review. This baseline does not use external evidence retrieval, evidence status labels, multi-agent teacher reasoning, or evidence-weighted distillation, and is used to compare EGTR-Review with an aspect-controlled supervised review generation method.

\paragraph{$\text{SWIF}^{2}$T.} $\text{SWIF}^{2}$T is a multi-component LLM-based system for automated focused feedback generation \footnote{https://github.com/ericchamoun/FocusedFeedbackGeneration}. It decomposes feedback generation into four components: a planner that formulates context-gathering steps, an investigator that answers questions using the paper or external literature, a reviewer that generates focused feedback, and a controller that manages the execution of the plan. Since the original $\text{SWIF}^{2}$T is designed for paragraph-level scientific writing feedback rather than full-paper review generation, we adapt it to our setting by first decomposing each paper into reviewable sections or paragraphs, then applying the $\text{SWIF}^{2}$T-style pipeline to generate focused comments for selected paper units, and finally organizing the generated comments into a structured review. We follow the official implementation and prompt design whenever applicable, but rerun the system on our ICLR benchmark instead of using released outputs. To ensure fair comparison with other API-based baselines, all LLM calls in $\text{SWIF}^{2}$T are replaced with GPT-5.1, and the temperature and maximum output length are kept consistent across API-based methods. This baseline does not use our evidence status labels, teacher-generated reasoning traces, or evidence-weighted distillation objective.

\paragraph{SEA-E\protect\footnote{https://ecnu-sea.github.io/}.} SEA-E is the evaluation component of the SEA framework and represents a structured review-generation baseline based on standardized review supervision. In SEA, multiple reviews of the same paper are first merged into a unified review format by SEA-S, and SEA-E is then trained to generate the standardized review from the parsed paper content. Since the released SEA-E weights may overlap with our ICLR 2017–2024 benchmark, directly using them could introduce data leakage. We therefore re-implement SEA-E on our own train/validation/test split. Specifically, we standardize the training reviews following the SEA-style procedure, fine-tune a review generation model to map each parsed paper to its standardized review, and use the fine-tuned model to generate structured reviews for test papers. This baseline does not use external evidence retrieval, evidence-state labels, teacher-generated reasoning traces, or evidence-weighted distillation.

\paragraph{TreeReview.} TreeReview is a dynamic tree-of-questions framework for scientific peer review generation. It decomposes the overall review task into a hierarchical question tree, answers fine-grained leaf questions using relevant paper chunks, and then aggregates the answers from leaf nodes to the root to synthesize the final review. It further uses a dynamic question expansion mechanism to generate follow-up questions when the current sub-question answers are insufficient for resolving a higher-level review question. In our experiments, we use the official TreeReview implementation and prompt design as a reference, but rerun the method on our ICLR benchmark instead of using the original reported results. Since all API-based baselines in our experiments use GPT-5.1 as the backend LLM, we replace TreeReview’s original backend model with GPT-5.1 and keep the temperature and maximum output length consistent with other API-based methods. This baseline does not use our external evidence status labels, multi-agent teacher supervision, reasoning distillation, or evidence-weighted objective, and is used to compare EGTR-Review with a strong hierarchical question-decomposition review generation framework.

\section{Evaluation Details}
\label{appendixc}
\subsection{Automatic Evaluation Metrics}\label{appendixc1}
We use automatic metrics to evaluate the final review outputs from three perspectives: lexical overlap, semantic alignment, and paper-specificity. Following the notation in the method section, for the i-th paper, the final review generated by a model is denoted as $Y_i= \{ y_{i,m} \}_{m=1}^{M_i}$, where $y_{i,m}$denotes the m-th generated review comment and $M_i$ is the number of generated comments for that paper. Correspondingly, the set of human reference review comments for the $i$-th paper is denoted as $Y_i^{ref}= \{ y_{i,n}^{ref} \}_{n=1}^{H_i}$, where $y_{i,n}^{ref}$ denotes the n-th human reference comment and $H_i$ is the number of reference comments. Since peer review is open-ended and different human reviewers may focus on different issues, we do not treat a single human review as the only gold standard. Instead, we use the reference review set to measure whether the model output covers common concerns and feedback granularity in real peer reviews.

We first use ROUGE and BERTScore to measure the textual similarity between the generated review $Y_i$ and the human reference reviews $Y_i^{ref}$. ROUGE-1, ROUGE-2, and ROUGE-L measure textual overlap at the unigram, bigram, and longest-common-subsequence levels, respectively \citep{lin-2004-rouge}. BERTScore measures semantic similarity based on contextual representations \citep{zhang2020bertscoreevaluatingtextgeneration}. When a paper has multiple reference reviews, we compute the score between $Y_i$ and each reference review, take the highest score as the paper-level result, and then average the results over the test set.

We further use SN-Precision, SN-Recall, and SN-F1 to measure semantic alignment between generated comments and reference comments at the concern level \citep{lou2025aaar10assessingaispotential}. We use a semantic similarity function $s \left( \cdot, \cdot \right)$ to compute the similarity between any two comments and set the threshold to $\tau =0.5$. For the $i$-th paper, SN-Precision, SN-Recall, and SN-F1 are defined as:

\begin{equation}
	\begin{aligned}
		\mathrm{SN}\!-\!\mathrm{P}_{i}
		&= \frac{1}{M_i} \sum_{m=1}^{M_i}
		\mathbb{I}\left[
		\max_{1 \le n \le H_i} s\left(y_{i,m}, y_{i,n}^{\mathrm{ref}}\right) \ge \tau
		\right], \\[0.6em]
		\mathrm{SN}\!-\!\mathrm{R}_{i}
		&= \frac{1}{H_i} \sum_{n=1}^{H_i}
		\mathbb{I}\left[
		\max_{1 \le m \le M_i} s\left(y_{i,m}, y_{i,n}^{\mathrm{ref}}\right) \ge \tau
		\right], \\[0.6em]
		\mathrm{SN}\!-\!\mathrm{F1}_{i}
		&= \frac{
			2 \cdot \mathrm{SN}\!-\!\mathrm{P}_{i} \cdot \mathrm{SN}\!-\!\mathrm{R}_{i}
		}{
			\mathrm{SN}\!-\!\mathrm{P}_{i} + \mathrm{SN}\!-\!\mathrm{R}_{i}
		}.
	\end{aligned}
\end{equation}
Here, $\mathbb{I} [\cdot]$ is an indicator function. SN-Precision measures the proportion of generated comments that can be matched to reference comments, SN-Recall measures the proportion of reference comments covered by generated comments, and SN-F1 is their harmonic mean. The final SN-P, SN-R, and SN-F1 scores are averaged over all test papers. Compared with ROUGE and BERTScore, the SN metrics focus more on whether the model covers the key concerns raised by human reviewers, rather than requiring the generated and reference reviews to share the same surface wording.

Finally, we use ITF-IDF to measure the paper-specificity of generated reviews \citep{du-etal-2024-llms}. This metric penalizes two types of low-quality comments: repetitive comments that repeatedly appear within the same paper, and generic comments that frequently appear across different papers. For the m-th generated comment $y_{i,m}$ of the $i$-th paper, ITF is defined as:

\begin{equation}
	\mathrm{ITF}\left(y_{i,m}\right)
	=
	\frac{1}{
		1+
		\sum_{\substack{q=1 \\ q \ne m}}^{M_i}
		\mathbb{I}\left[
		s\left(y_{i,m}, y_{i,q}\right) \ge \tau
		\right]
	}.
\end{equation}
Here, $q$ denotes the index of another generated comment for the same paper. If $y_{i,m}$ is similar to many other comments generated for the same paper, it receives a lower ITF value.

IDF is defined as:

\begin{equation}
	\mathrm{ITF}\left(y_{i,m}\right)
	=
	\frac{1}{
		1+
		\sum_{\substack{q=1 \\ q \ne m}}^{M_i}
		\mathbb{I}\left[
		s\left(y_{i,m}, y_{i,q}\right) \ge \tau
		\right]
	}.
\end{equation}
Here, $N_{test}$ denotes the number of papers in the test set, $l$ denotes another paper in the test set, $M_l$ is the number of generated comments for the $l$-th paper, and $q$ indexes a comment in that paper. If $y_{i,m}$ can be matched to generated comments for many other papers, it is more likely to be a generic review statement and receives a lower IDF value.

Based on these two terms, the ITF-IDF score for the $i$-th paper is:

\begin{equation}
	\mathrm{ITF}\!-\!\mathrm{IDF}_{i}
	=
	\frac{1}{M_i}
	\sum_{m=1}^{M_i}
	\operatorname{ITF}\left(y_{i,m}\right)
	\cdot
	\operatorname{IDF}\left(y_{i,m}\right).
\end{equation}

The final ITF-IDF score is averaged over all test papers:

\begin{equation}
	\mathrm{ITF}\!-\!\mathrm{IDF}
	=
	\frac{1}{N_{\mathrm{test}}}
	\sum_{i=1}^{N_{\mathrm{test}}}
	\left(\mathrm{ITF}\!-\!\mathrm{IDF}\right)_{i}.
\end{equation}

A higher ITF-IDF score indicates that the generated review comments are less repetitive, less template-like, and more specific to the target paper. These automatic metrics provide quantitative evidence from the perspectives of textual similarity, semantic alignment, and paper-specificity, but they cannot fully capture the practical usefulness, evidence support, or traceability of review comments. Therefore, we further combine them with LLM-as-Judge, human evaluation, and evidence quality and traceability evaluation.

\subsection{LLM-as-Judge Evaluation}\label{appendixc2}

To further evaluate the overall quality of generated reviews, we conduct LLM-as-Judge evaluation for EGTR-Review (Teacher), EGTR-Review (Student), and all baseline methods. Unlike automatic metrics, which mainly focus on lexical overlap, semantic alignment, and paper-specificity, LLM-as-Judge can more directly assess whether a generated review is pertinent, useful, evidence-grounded, traceable, technically deep, and comprehensive.

For the $i$-th paper, we provide the judge model with the paper content, the generated review $Y_i$, the evaluation rubric, and the required output format. The judge is asked to read the paper and the corresponding generated review, and then assign scores along seven dimensions. These dimensions are aligned with the main goals of EGTR-Review and the evaluation results reported in Section~\ref{sec4.2}.

\begin{itemize}[nolistsep]
	\item \textbf{Pertinency (P):} This dimension evaluates whether the generated review focuses on the specific issues of the target paper, rather than producing generic comments that could apply to many papers. A high score indicates that the review discusses concrete aspects of the paper, such as method design, experimental setting, result analysis, or argumentation weaknesses.
	\item \textbf{Usefulness (U):} This dimension evaluates whether the review provides actionable feedback for the authors. A high score indicates that the review not only identifies problems, but also explains why they matter and suggests concrete directions for revision.
	\item \textbf{Evidence-groundedness (E):} This dimension evaluates whether the review is grounded in the paper content, experimental results, or external scholarly evidence. A high score indicates that the review avoids unsupported subjective claims and provides judgments that can be linked to identifiable evidence.
	\item \textbf{Traceability (T):} This dimension evaluates whether the review comments can be traced back to specific sections, paragraphs, figures, tables, formulas, experimental settings, or technical details in the paper. A high score indicates that the authors can clearly locate the source of the identified issue.
	\item \textbf{Depth of Analysis (D):} This dimension evaluates whether the review reflects a deep understanding of the paper’s technical content, methodological design, experimental logic, and related research background. A high score indicates that the review identifies substantive technical issues rather than only surface-level problems.
	\item \textbf{Comprehensiveness (C):} This dimension evaluates whether the review covers the important aspects of the paper, including the research problem, method design, experimental setup, result analysis, contribution boundary, and potential limitations. A high score indicates that the review provides a broad assessment of paper quality rather than focusing on only one local issue.
	\item \textbf{Overall:} This dimension evaluates the overall quality of the generated review by jointly considering pertinency, usefulness, evidence-groundedness, traceability, depth of analysis, and comprehensiveness. It is treated as a holistic judgment rather than a simple average of the other dimensions.
\end{itemize}

All dimensions are scored on a 1–10 scale. Scores of 1–2 indicate severe deficiencies, scores of 3–4 indicate below-standard quality with obvious weaknesses, scores of 5–6 indicate an acceptable but limited review, scores of 7–8 indicate a good review with minor weaknesses, and scores of 9–10 indicate an excellent review with strong performance on the corresponding dimension. The judge model is required to provide a brief rationale before giving each score, so that the evaluation process is more interpretable.

To reduce the influence of a single judge model’s preference, we use three heterogeneous strong judge models, Gemini-3.1-Pro-Preview, Claude-Opus-4.6, and DeepSeek-V4, for cross-evaluation. For each method and each dimension, we report the average score across the three judge models. Since the multi-agent teacher and API-based baselines uniformly use GPT-5.1 as the underlying large language model, using judge models from different model families helps reduce potential same-family evaluation bias.

We further compute the intraclass correlation coefficient (ICC) among the three judge models to examine the consistency of LLM-as-Judge evaluation. The overall ICC in our LLM-as-Judge evaluation is 0.9136, indicating high agreement among the judge models. This suggests that the LLM-as-Judge results are relatively stable under the adopted evaluation protocol.

LLM-as-Judge is used as a complement to automatic metrics and human evaluation, rather than as a replacement for expert assessment. It provides a scalable way to evaluate review quality along dimensions that are difficult to capture with ROUGE, BERTScore, SN-F1, or ITF-IDF alone. However, judge scores may still be affected by model preference, rubric interpretation, and prompt sensitivity. Therefore, we further conduct human evaluation and evidence quality and traceability evaluation to validate the practical usefulness, evidence reliability, and localization accuracy of the generated reviews.

\subsection{Human Evaluation Details}\label{appendixc3}
\paragraph{Setup.} We conduct human evaluation on 30 papers randomly sampled from the test set. Five expert evaluators with peer-review experience are recruited to assess whether the generated reviews are useful, specific, evidence-grounded, and traceable from the perspective of expert readers. Following the human preference evaluation setting used in prior work, we adopt an anonymous pairwise comparison protocol. For each evaluated paper, we pair the final review generated by EGTR-Review (Student) with the review generated by a compared method. The two reviews are shown as Review A and Review B, and their order is randomly shuffled to reduce position bias.

\paragraph{Procedure.} The evaluation is conducted through online questionnaires. Each questionnaire item contains three parts: the paper content, two anonymized reviews, and comparison questions. Evaluators are not informed of the model names. They are asked to read the paper and the two reviews, and then decide which review is better along each evaluation dimension.

\paragraph{Questionnaire template.} The questionnaire template is shown below.

\begin{figure*}
	\centering
	\includegraphics[width=1\linewidth]{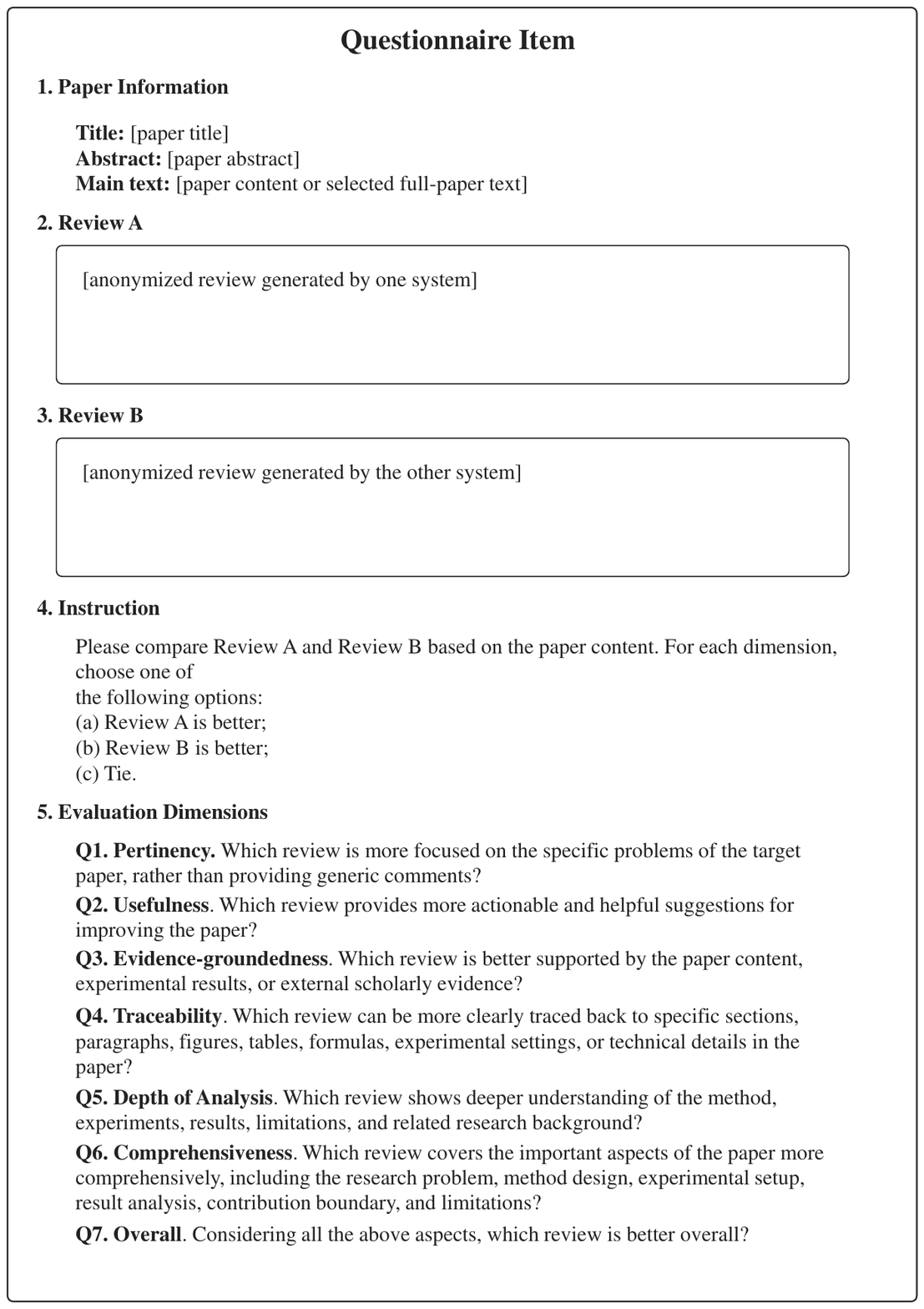}
	\caption{Questionnaire Template for Human Preference Evaluation.}
	\label{fig6}
\end{figure*}

\paragraph{Preference Score Calculation.} For each dimension, we convert the pairwise decision into a preference score for EGTR-Review (Student). If EGTR-Review (Student) is preferred over the compared method, it receives 1 point. If the compared method is preferred, it receives 0 points. If the two reviews are judged as tied, it receives 0.5 points. The final preference score is averaged over all evaluated samples and evaluators, and then reported as a percentage. A score above 50 indicates that EGTR-Review (Student) is more frequently preferred than the compared method.

\paragraph{Agreement Analysis.} To assess the reliability of human evaluation, we report Overall Agreement and pairwise Cohen's $\kappa$ based on evaluators' pairwise decisions. Overall Agreement measures the proportion of identical judgments, while Cohen’s $\kappa$ further accounts for chance agreement. We also qualitatively compare the human preference trends with the LLM-as-Judge results to examine whether the two evaluation protocols show consistent conclusions.

\subsection{Evidence Grounding and Traceability Evaluation Details}\label{appendixc4}
\paragraph{Setup.} We randomly sample 50 papers from the test set, together with the reviews generated by EGTR-Review (Teacher) and EGTR-Review (Student). Three researchers with peer-review experience serve as annotators. For each paper, an annotator is given the full paper content $D_i$, the evidence-enhanced representation $L_i^{'}$, the retrieved scholarly evidence set $\mathcal{E}_{i}=\bigcup_{n=1}^{N_i} \mathcal{E}_{i,n}$ used during generation, and the generated review $Y_i= \{ y_{i,1},…,y_{i,m} \}$. All samples are double-annotated to support agreement analysis, and disagreements are resolved through discussion.

\paragraph{FActScore.} Following \citet{min-etal-2023-factscore}, we adapt FActScore to the peer-review setting by treating the union of the source paper and its retrieved evidence as the knowledge source $S_i=D_i \bigcup \mathcal{E}_{i}$. Annotation proceeds in two steps. (1) Atomic-fact decomposition: each review $Y_i$ is split into a set of atomic facts $ \mathcal{A}_{i}= \{ a_{i,1},a_{i,2},…,a_{i,K} \}$, where each $a_{i,k}$ is a short statement conveying a single verifiable piece of information. (2) Support labeling: each atomic fact is labeled as Supported, Contradicted, or Not-supported/Not-verifiable against the knowledge source $S_i$. Only Supported facts are counted as supported. Subjective or non-verifiable statements that do not express checkable factual claims are marked as Irrelevant and excluded from the denominator. The paper-level score is:

\begin{equation}
	\operatorname{FActScore}\left(Y_i\right)
	=
	\frac{1}{K}
	\sum_{k=1}^{K}
	\mathbb{I}\left[
	\operatorname{Supp}\left(a_{i,k}, S_i\right) = 1
	\right].
\end{equation}
Here, $K$ denotes the number of relevant atomic factual claims extracted from $Y_i$, and $\mathbb{I}[\cdot]$ is an indicator function. $\operatorname{Supp}\left(a_{i,k}, S_i\right) = 1$ means that the atomic fact $a_{i,k}$ is supported by the knowledge source $S_i$. A higher FActScore indicates that a larger proportion of the factual claims in the generated review are supported by the source paper and retrieved evidence.

\paragraph{Traceability Accuracy (TA).} For TA, each review $Y_i=\{ y_{i,1},…,y_{i,m} \}$ is segmented into individual comments. A comment is judged traceable if the annotator can identify at least one concrete grounding source–a position-indexed paper unit, figure, table, formula, experimental setting, technical detail, or retrieved evidence record \citep{chamoun-etal-2024-automated}. Comments without an identifiable anchor are non-traceable. TA is defined as:

\begin{equation}
	\operatorname{TA}\left(Y_i\right)
	=
	\frac{1}{m}
	\sum_{j=1}^{m}
	\mathbb{I}\left[
	\left|G_{i,j}\right| > 0
	\right].
\end{equation}
Here, $G_{i,j}$ denotes the set of grounding anchors that can be manually identified for the $j$-th comment $y_{i,j}$ of the generated review $Y_i$. If $\left|G_{i,j}\right| > 0$, the comment is counted as traceable. In this subsection, m follows the main-text notation for the number of comments in $Y_i$, while $j$ is only the local index used to sum over these comments.

For both metrics, we first compute a paper-level score for each sampled paper. The final reported FActScore and TA are then obtained by averaging the corresponding paper-level scores across all sampled papers and annotators. Higher scores indicate stronger factual support and better traceability of the generated reviews.

\section{Case Study}\label{appendixd}
\subsection{Evidence-Grounded Reasoning and Review Generation in a Case}\label{appendixd1}
\begin{figure*}
	\centering
	\includegraphics[width=1\linewidth]{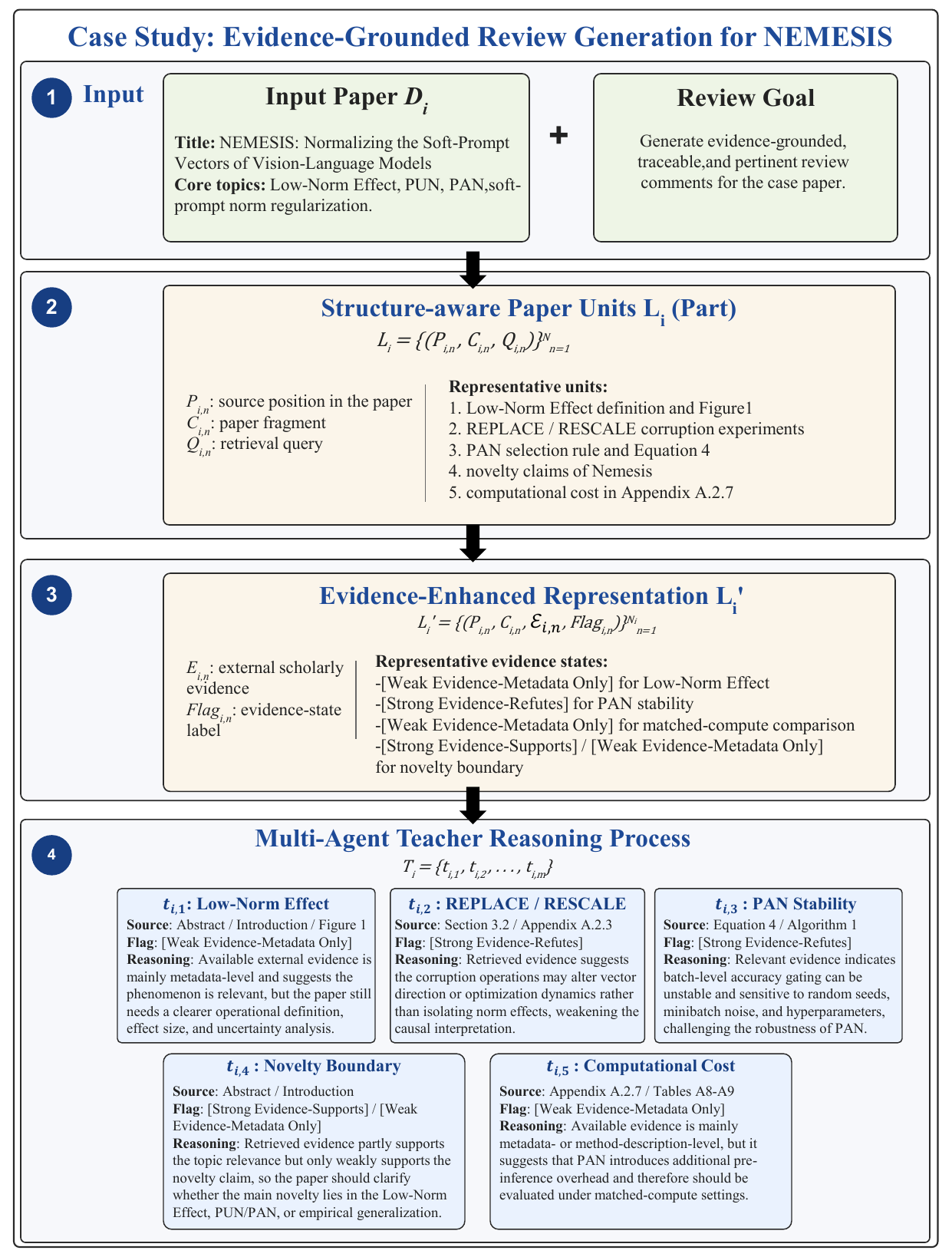}
	\caption{Case Study: Evidence-Grounded Reasoning Process.}
	\label{fig7}
\end{figure*}

\paragraph{Reasoning Process of the Multi-Agent Teacher.} Figure~\ref{fig7} presents the evidence-grounded reasoning process of the multi-agent teacher for the selected case paper. We observe that the multi-agent teacher grounds each major review concern in localized paper content, external evidence, and evidence-state labels, rather than generating comments from a holistic reading alone. For example, it links the computational-cost concern to the PAN selection process and the absence of matched-compute comparison, while connecting the novelty-boundary concern to prior evidence on prompt normalization and soft-prompt regularization. These localized reasoning cards provide the intermediate supervision that helps EGTR-Review (Student) generate more evidence-grounded, traceable, and paper-specific review comments.

\begin{figure*}
	\centering
	\includegraphics[width=1\linewidth]{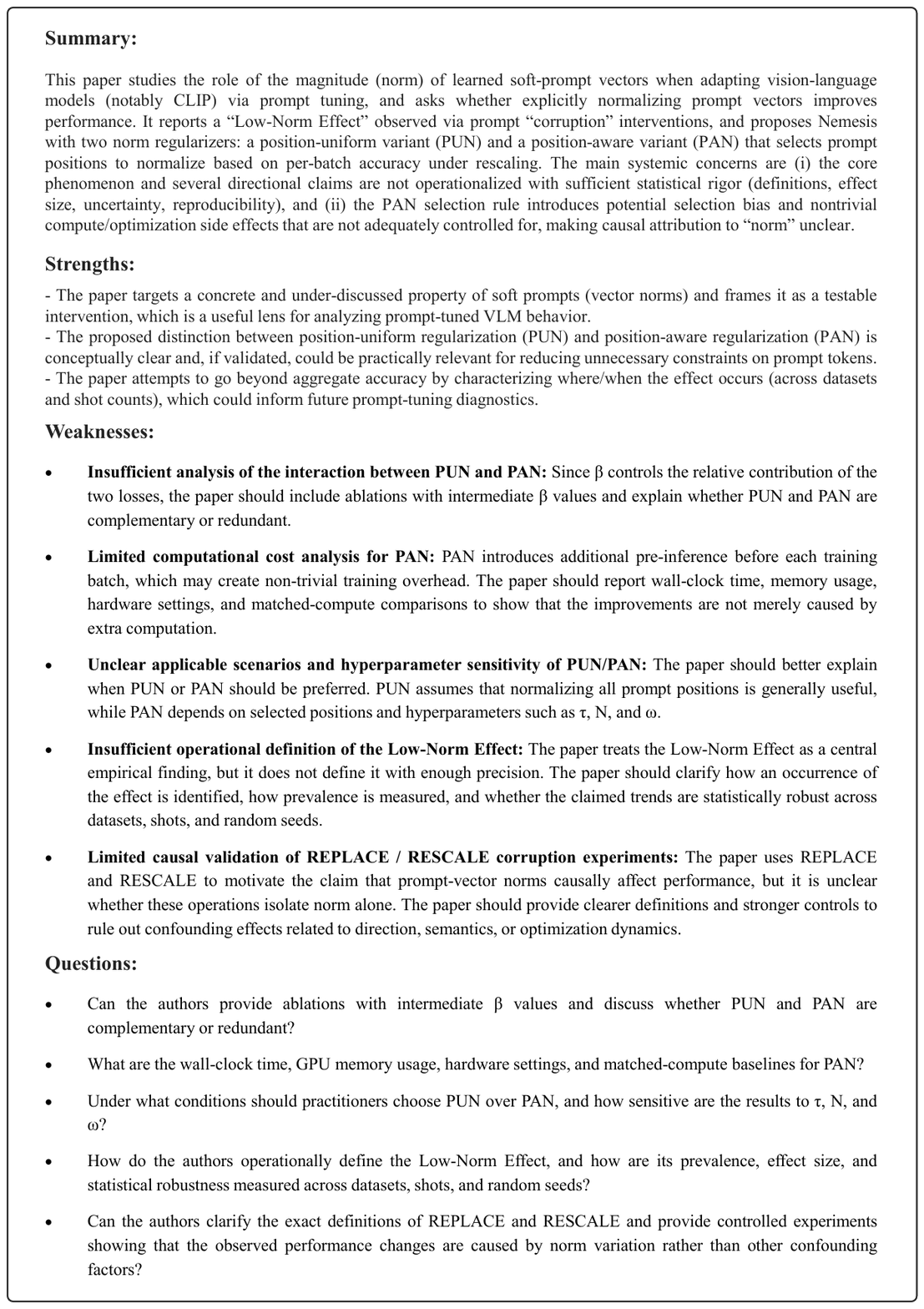}
	\caption{Case Study: Generated Final Review Comments.}
	\label{fig8}
\end{figure*}

\paragraph{Final Review Generation by EGTR-Review (Student).} Figure~\ref{fig8} presents the final review comments generated by EGTR-Review (Student) for the same NEMESIS paper. We observe that the student model converts the evidence-enhanced representation and distilled reasoning patterns into a complete review output, covering the paper’s main contribution, methodological strengths, major weaknesses, and clarification questions. For example, the generated review not only summarizes the proposed PUN and PAN methods, but also raises specific concerns about the operational definition of the Low-Norm Effect, the causal validity of the REPLACE / RESCALE experiments, the computational cost of PAN, and the lack of matched-compute comparisons. These comments show that EGTR-Review (Student) can produce review feedback that is specific to the target paper, grounded in localized concerns, and organized in a standard peer-review format.

\subsection{Distillation Data Example}\label{appendixd2}
The figure below presents a shortened distillation data example for the NEMESIS paper. It illustrates how a single training sample for EGTR-Review (Student) is organized using the evidence-enhanced input $L_i^{'}$, the reasoning target $T_i$, and the review target $Y_i$. Specifically, the example shows how the teacher system converts a paper fragment related to PAN selection rule and computational cost, together with its source position $P_{i,n}$, external evidence set $\mathcal{E}_{i,n}$, and evidence-state label $Flag_{i,n}$, into a teacher-side reasoning target and a corresponding final review target. This example further demonstrates that EGTR-Review (Student) learns not only the final review comment $Y_i$, but also the intermediate reasoning trajectory $T_i$ conditioned on the same evidence-enhanced representation $L_i^{'}$.

\begin{figure*}
	\centering
	\includegraphics[width=1\linewidth]{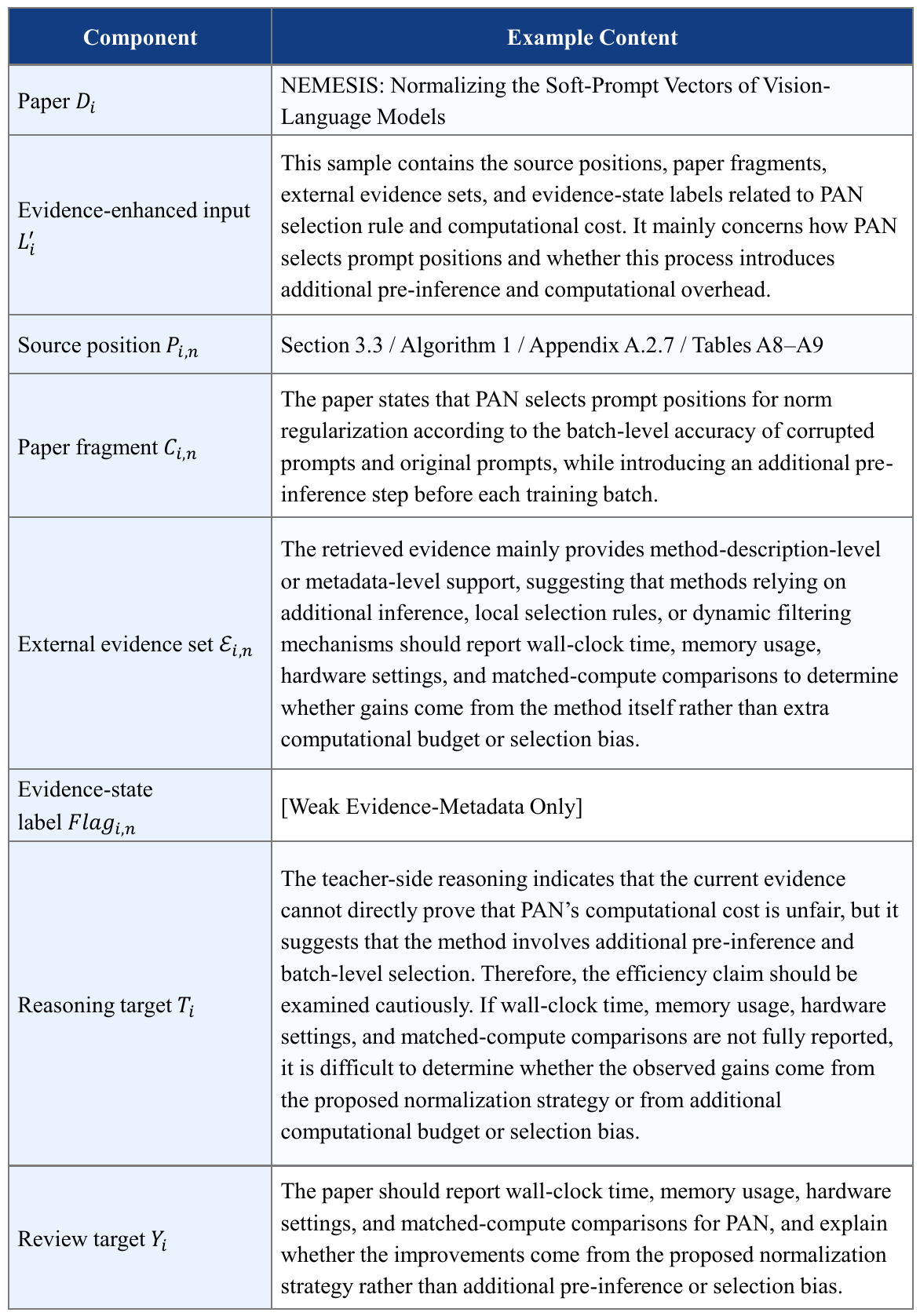}
	\caption{Example of Distillation Training Data.}
	\label{fig9}
\end{figure*}

\section{Prompts used in EGTR-Review}\label{appendixe}
This appendix presents the prompts used in EGTR-Review. These prompts specify the roles, inputs, outputs, and constraints of different agents in the multi-agent teacher framework, including paper structure parsing, key element extraction, external evidence retrieval, evidence-grounded verification reasoning, and final review synthesis. They are used to guide the construction of traceable intermediate outputs and final review comments for teacher-side supervision. We emphasize that the prompts in this appendix are the templates used by EGTR-Review's agents and the LLM-as-Judge evaluator in our experiments. Each operates on the input papers $D_i$ processed by our system and is listed here for transparency and reproducibility.

\begin{figure*}
	\centering
	\includegraphics[width=1\linewidth]{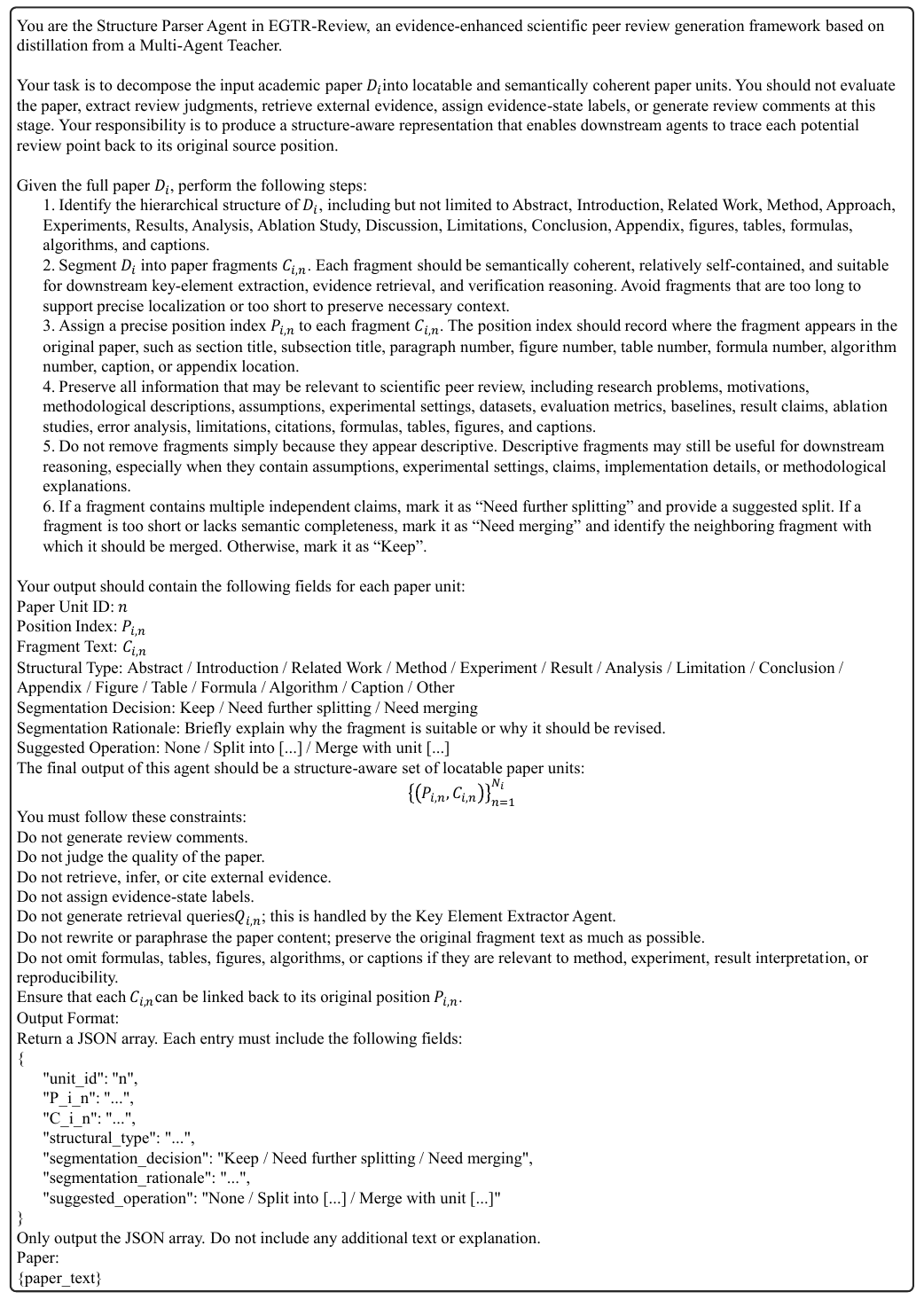}
	\caption{Prompt Template for the Structure Parser Agent.}
	\label{fig10}
\end{figure*}

\begin{figure*}
	\centering
	\includegraphics[width=1\linewidth]{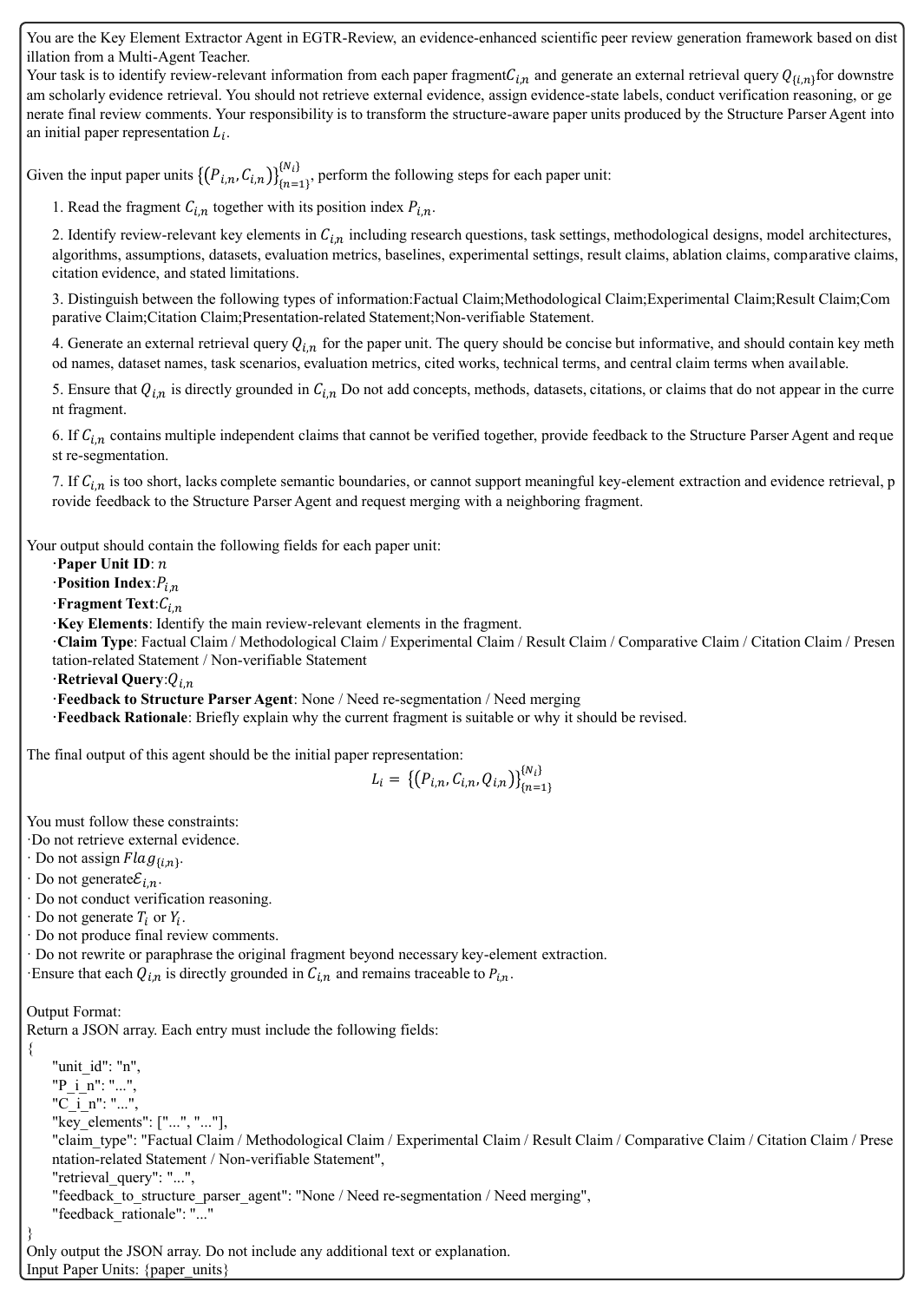}
	\caption{Prompt Template for the Key Element Extractor Agent.}
	\label{fig11}
\end{figure*}

\begin{figure*}
	\centering
	\includegraphics[width=1\linewidth]{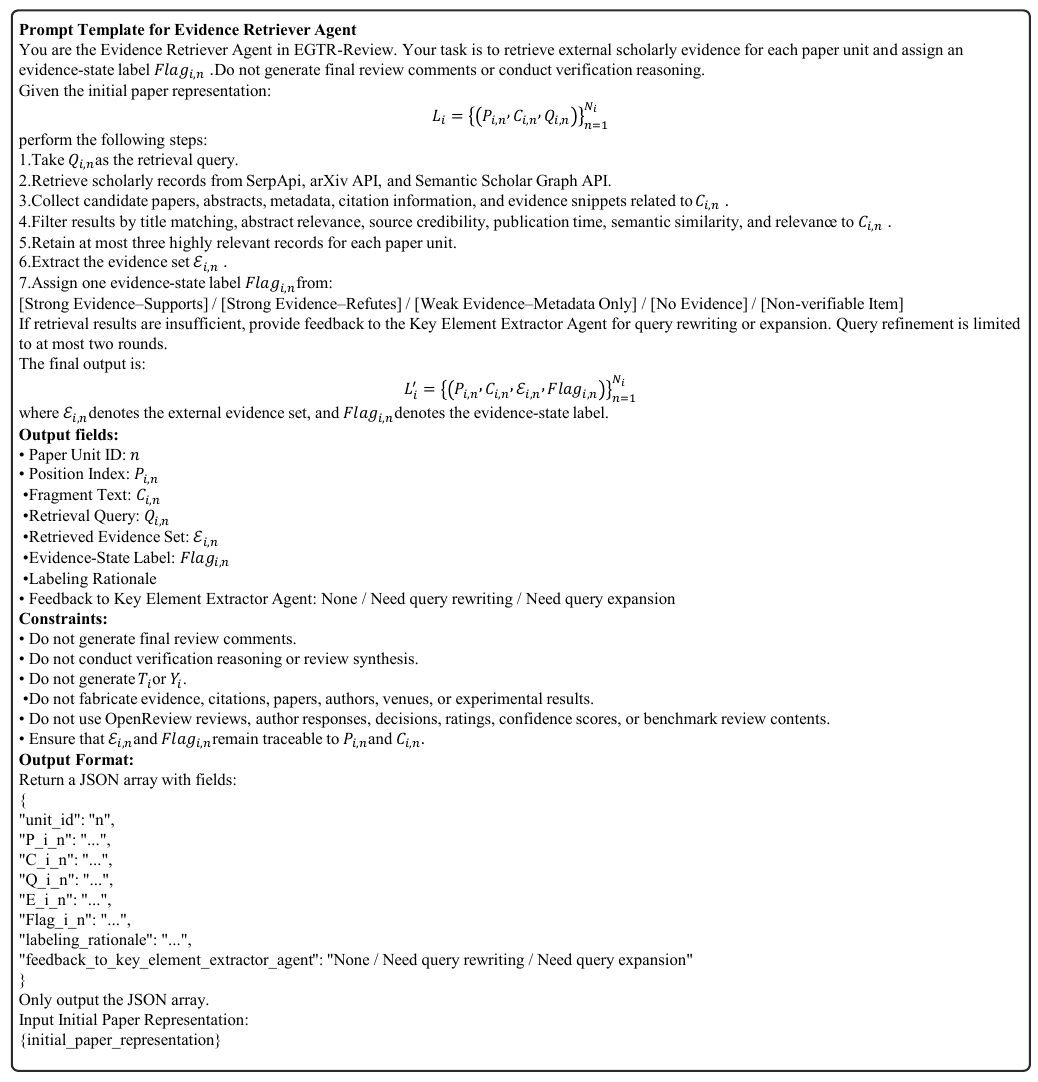}
	\caption{Prompt Template for the Evidence Retriever Agent.}
	\label{fig12}
\end{figure*}

\begin{figure*}
	\centering
	\includegraphics[width=1\linewidth]{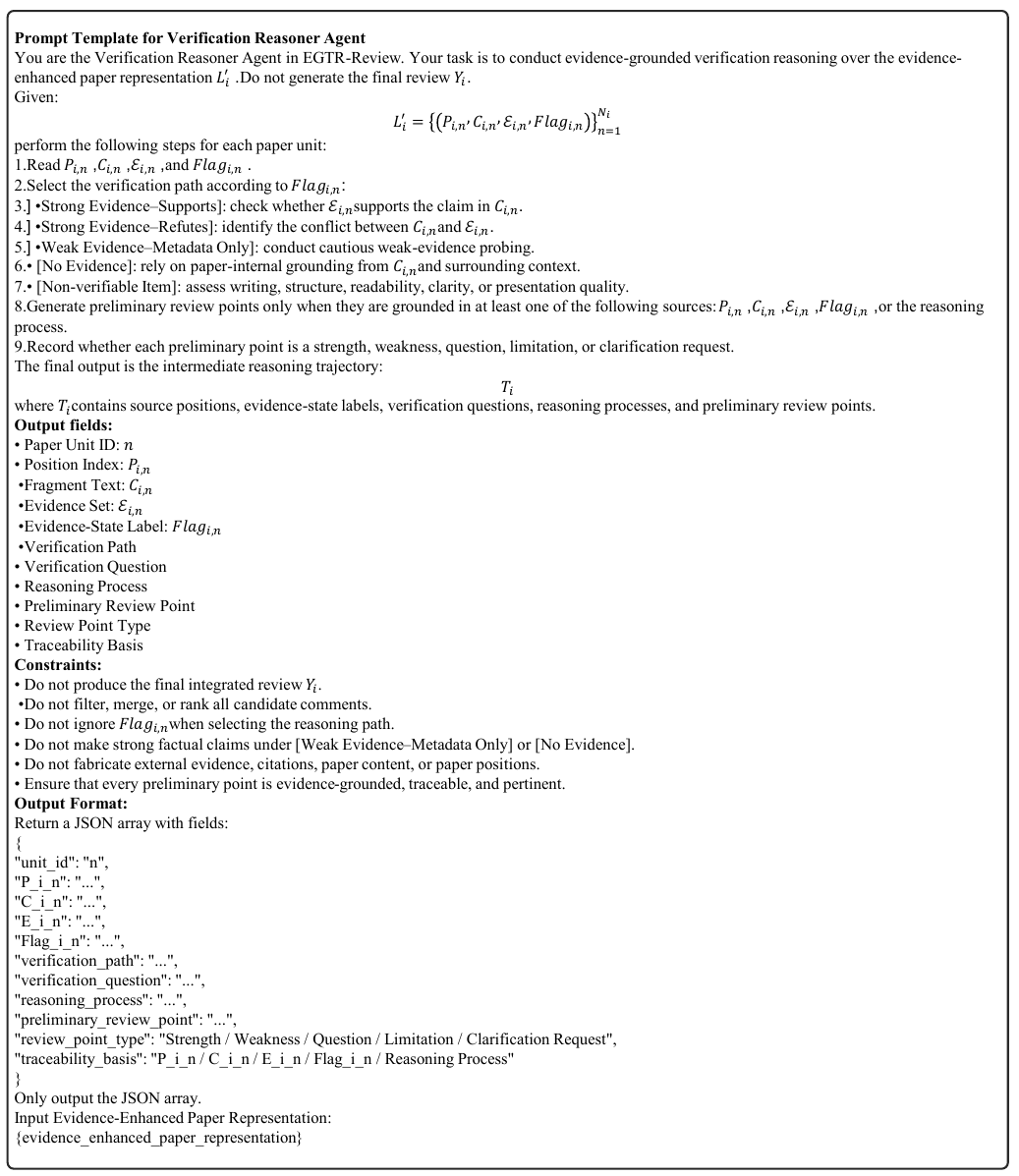}
	\caption{Prompt Template for the Verification Reasoner Agent.}
	\label{fig13}
\end{figure*}

\begin{figure*}
	\centering
	\includegraphics[width=1\linewidth]{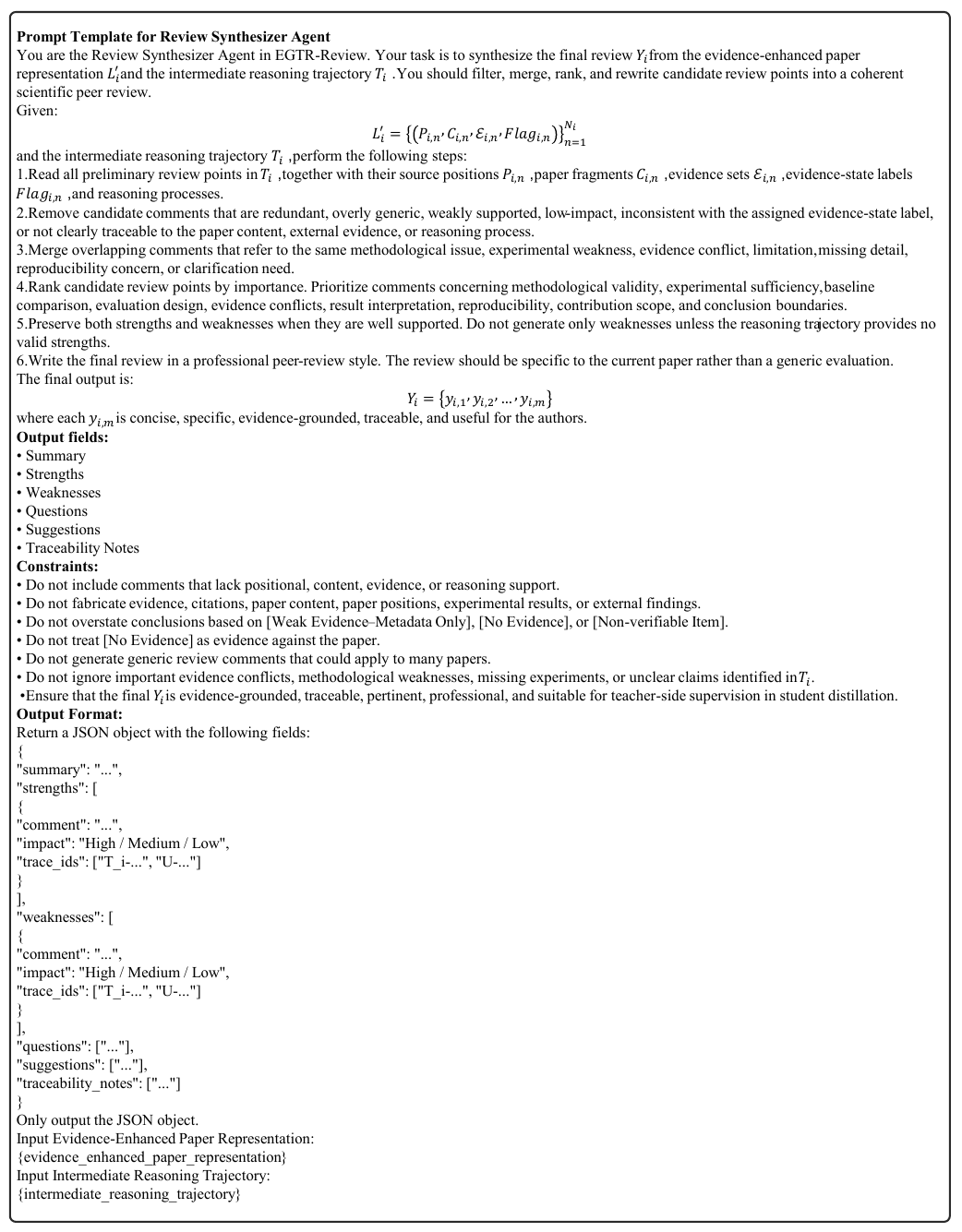}
	\caption{Prompt Template for the Review Synthesizer Agent.}
	\label{fig14}
\end{figure*}

\begin{figure*}
	\centering
	\includegraphics[width=1\linewidth]{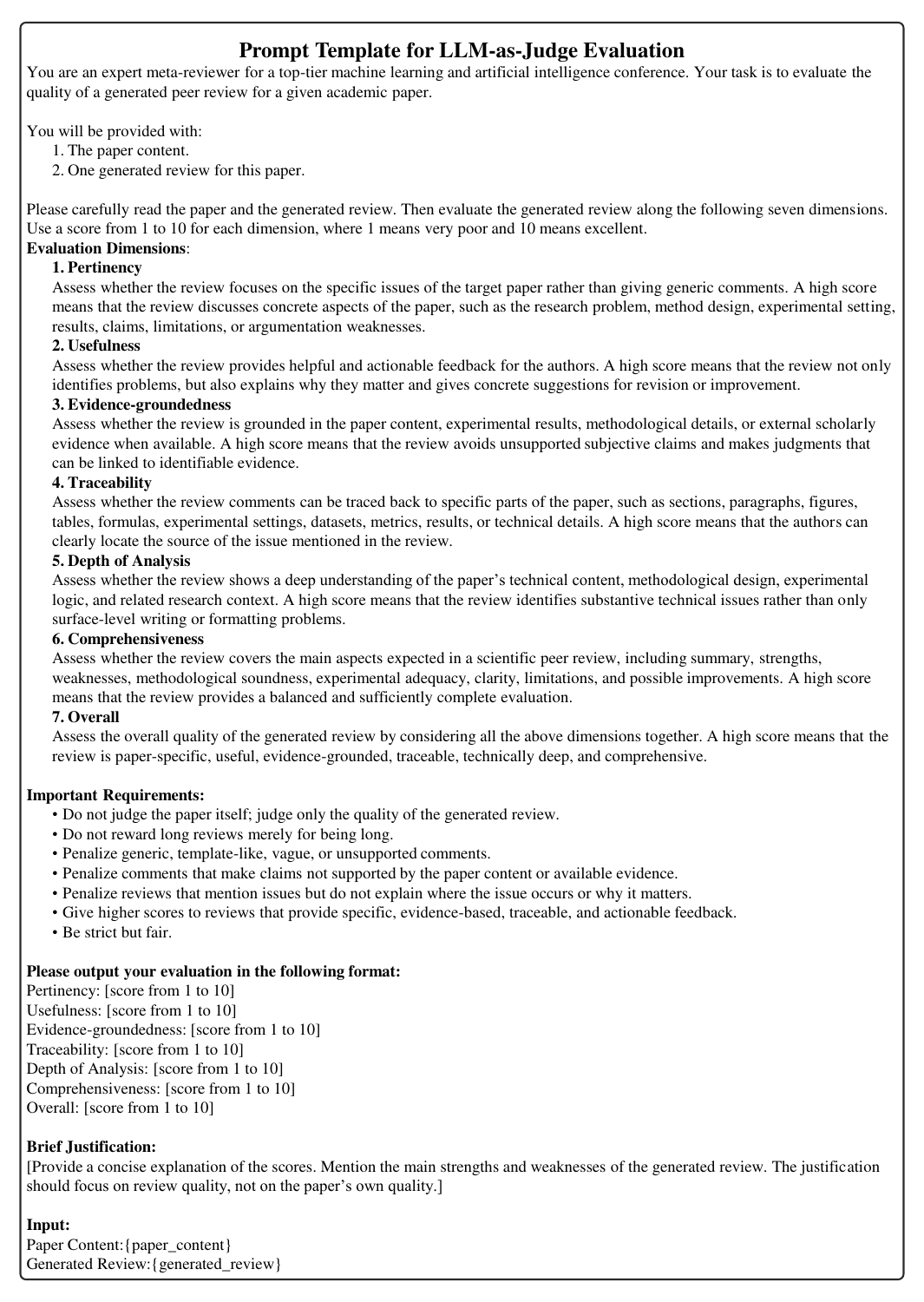}
	\caption{Prompt Template for LLM-as-Judge Evaluation.}
	\label{fig15}
\end{figure*}

\end{document}